%% file: main.tex
\documentclass[sn-mathphys]{sn-jnl}

\jyear{2021}%

\theoremstyle{thmstyleone}%
\newtheorem{theorem}{Theorem}
\newtheorem{proposition}[theorem]{Proposition}%
\newtheorem{example}{Example}%

\theoremstyle{thmstyletwo}%

\theoremstyle{thmstylethree}%

\usepackage{amsfonts}       
\usepackage{amsmath}
\usepackage{amsthm}
\usepackage{nicefrac}       
\usepackage{xfrac}
\usepackage{caption}
\usepackage{multirow}
\usepackage{booktabs}       
\usepackage{diagbox}
\usepackage{xcolor}
\usepackage{algorithm}
\usepackage{algpseudocode}
\usepackage{varwidth}
\usepackage{enumitem}

\algdef{SE}{WithProb}{EndWithProb}[1]{\textbf{with probability} \(\mbox{#1}\) \textbf{do}}{\textbf{end}}

\newtheorem{lemma}{Lemma}

\DeclareMathOperator*{\argmin}{arg\,min}

\begin{document}

\title[Domain Adversarial Neural Networks for Domain Generalization]{Domain Adversarial Neural Networks for Domain Generalization: When It Works and How to Improve}


\author[1]{\fnm{Anthony} \sur{Sicilia}}\email{anthonysicilia@pitt.edu}

\author[2]{\fnm{Xingchen} \sur{Zhao}}\email{zhao.xingc@northeastern.edu}

\author*[3]{\fnm{Seong Jae} \sur{Hwang}}\email{seongjae@yonsei.ac.kr}




\affil*[1]{\orgdiv{Intelligent Systems Program}, \orgname{University of Pittsburgh}, \orgaddress{\street{210 S. Bouquet Street}, \city{Pittsburgh}, \postcode{15260}, \state{PA}, \country{USA}}}

\affil[2]{\orgdiv{Department of Electrical \& Computer Engineering}, \orgname{Northeastern University}, \orgaddress{\street{360 Huntington Ave}, \city{Boston}, \postcode{02115}, \state{MA}, \country{USA}}}

\affil[3]{\orgdiv{Department of Artificial Intelligence}, \orgname{Yonsei University}, \country{Korea}}

\abstract{\input{0_abstract}}




\maketitle

\section{Introduction}
\label{sec:intro}
\input{1_intro}

\section{Domain Adversarial Neural Network (DANN)}
\label{sec:dann}
\input{2_dann}

\section{Understanding Domain Alignment in Domain Generalization}
\label{sec:ours_theory}
\input{3_ours_theory}

\section{An Algorithmic Extension to DANN}
\label{sec:ours_algorithm}
\input{4_ours_algorithm}

\section{Experimentation}
\label{sec:results}
\input{5_results}

\section{Related Works}
\label{sec:related}
\input{6_related}

\section{Discussion}
\label{sec:discussion}
\input{7_conclusion}

\clearpage

\begin{appendices}

\section{}
\label{sec:appendix_a}
\input{8_appendix_a}

\clearpage
\section{}
\label{sec:appendix_b}
\input{9_appendix_b}

\end{appendices}


\clearpage

\bibliography{main}


\end{document}

%% file: 0_abstract.tex
Theoretically, domain adaptation is a well-researched problem. Further, this theory has been well-used in practice. In particular, we note the bound on target error given by Ben-David et al. (2010) and the well-known domain-aligning algorithm based on this work using Domain Adversarial Neural Networks (DANN) presented by Ganin and Lempitsky (2015). Recently, multiple variants of DANN have been proposed for the related problem of domain \textit{generalization}, but without much discussion of the original motivating bound. In this paper, we investigate the validity of DANN in domain generalization from this perspective. We investigate conditions under which application of DANN makes sense and further consider DANN as a dynamic process during training. Our investigation suggests that the application of DANN to domain generalization may not be as straightforward as it seems. To address this, we design an algorithmic extension to DANN in the domain generalization case. Our experimentation validates both theory and algorithm.   

%% file: 1_intro.tex
In general, in machine learning, we assume the training data for our learning algorithm is well representative of the testing data. That is, we assume our training data follows the same distribution as our testing data. Of primary interest to this paper is the case where this assumption fails to hold:
we consider learning in the presence of multiple \textit{domains}. We formalize the multiple domain problem of interest as the case where (at train-time) we observe $k$ domains referred to as \textit{sources} which have distributions $\mathbb{P}_1, \mathbb{P}_2, \ldots, \mathbb{P}_k$ over some space $\mathcal{X}$. At test-time, we are evaluated on a distinct \textit{target} domain which has distribution $\mathbb{Q}$ over $\mathcal{X}$. All of these feature distributions have (potentially) distinct labelling functions
and our goal is to learn the labeling function on the target. Typically, we assume some restriction on observation of the target domain at train-time. In the literature, a large amount of work is concerned with the problem of \textit{Domain Adaptation} (DA) which assumes access to samples from $\mathbb{Q}$, but restricts access to the labels of these samples. More recently, there has also been an active investigation on the problem of \textit{Domain Generalization} (DG) which instead assumes absolutely \textit{no} access to the target domain. In spite of these restrictions, in both cases, the goal is for our learning algorithm trained on \textit{sources} to perform well when evaluated on the \textit{target}.

One popular approach to DA is the use of a Domain Adversarial Neural Network (DANN) originally proposed by Ganin and Lempitsky \cite{ganin2015unsupervised}. Intuitively, this approach attempts to align the source and target domains by learning feature representations of both which are indiscernible by a domain discriminator trained to distinguish between the two distributions. Informally speaking, this seems like a sensible approach to DA. By accomplishing this \textit{domain alignment}, the neural network should still be adept at the learned task when it is evaluated on the target domain at test-time. While DANN was originally proposed for DA, the adoption of this reasoning has motivated adaptations of this approach for DG \cite{matsuura2019domain, albuquerque2019adversarial, li2018deep_conditional, li2018domain_mmd}. In fact, very early works in DG \cite{muandet2013domain} are similarly motivated by the goal of domain-agnostic feature representations.  

Still, it is worth noting that 
the original proposal of DANN \cite{ganin2015unsupervised} was motivated by theory. In particular, Ganin and Lempitsky base their algorithm on the target-error bound given by Ben-David et al. \cite{ben2007analysis, ben2010theory}. Under appropriate assumptions, interpretation of the bound suggests domain alignment as achieved through DANN should improve performance on the target distribution, but importantly, it motivates alignment between the source and target. Counter to this, DANN variants for DG generally align multiple source domains because no access to target data is permitted. This shortcoming gives rise to the question of primary interest to this paper:
\begin{quote}
    \textit{Is there a justification for source alignment using DANN in DG?}
\end{quote}

Specifically, we are concerned with a target-error bound similar to those provided by Ben-David et al. \cite{ben2010theory}. To answer this question, we appeal to a recent theoretical proposal by Albuquerque et al. \cite{albuquerque2019adversarial} which uses a reference object (i.e., the set of mixture distributions of the sources) to derive a target-error bound in the domain generalization setting. Building on this framework, we provide answers to two important considerations:
\begin{enumerate}
    \item What additional reference objects (besides sets of mixture distributions) satisfy the primary condition used to derive target-error bounds in DG?
    \item How does the target-error bound behave as a \textit{dynamic} quantity during the training process?
\end{enumerate}
Ultimately, answering these two questions allows us to formulate a novel extension of the Domain Adversarial Neural Network. We validate experimentally that this extension improves performance and otherwise agrees with our theoretical expectations.

%% file: 2_dann.tex
In this section, we cover the necessary background on Domain Adversarial Neural Networks (DANN). We first present the original bound on target-error in the case of unsupervised DA \cite{ben2007analysis, ben2010theory} which motivates the DANN algorithm proposed by Ganin and Lempitsky~\cite{ganin2015unsupervised}. Following this, we outline the key differences introduced by a DANN variant proposed by Matsuura and Harada~\cite{matsuura2019domain}. Although this variant achieves state-of-the-art (DANN) performance in DG, we point out the main concerns we have regarding the justification of this approach.

\subsection{In Domain Adaptation}
As mentioned, we begin with a motivating result of Ben-David et al. \cite{ben2010theory}. Intuitively, this result describes bounds on the target-error controlled, in part, by a computable measure of \textit{divergence} between distributions. While we provide a more detailed exposition of the problem setup in Appendix~\ref{sec:appendix_a}, we begin by listing here the key terms to familiarize the reader.

\paragraph{Setup} For a binary hypothesis $h$, a distribution $\mathbb{P}$, and a labeling function $f$ for $\mathbb{P}$, we define the \textit{error} $\mathcal{E}_\mathbb{P}(h)$ of $h$ on the distribution $\mathbb{P}$ as follows
\begin{equation}
\label{eqn:error}
    \mathcal{E}_\mathbb{P}(h) = \mathbf{E}_{x \sim \mathbb{P}} \left\lvert h(x) - f(x) \right \rvert = \mathbf{E}_{x \sim \mathbb{P}} \left [1[h(x) \neq f(x)]\right].
\end{equation}
This is our primary measure of the \textit{quality} of a hypothesis when predicting on a distribution $\mathbb{P}$. To measure differences in distribution, we use the $\mathcal{H}$-divergence which is an adaptation of the $\mathcal{A}$-distance \cite{kifer2004detecting}. In particular, given two distributions $\mathbb{P}$, $\mathbb{Q}$ over a space $\mathcal{X}$ and a corresponding hypothesis class $\mathcal{H} \subseteq \{h \mid h: \mathcal{X} \to \{0,1\}\}$, the $\mathcal{H}$-divergence \cite{ben2010theory} is defined  
\begin{equation}
    d_\mathcal{H}(\mathbb{P}, \mathbb{Q}) = 2 \sup_{h \in \mathcal{H}} \left\lvert \mathrm{Pr}_\mathbb{P}(I_h) - \mathrm{Pr}_\mathbb{Q}(I_h)\right\rvert
\end{equation}
where $I_h = \{x \in \mathcal{X} \mid h(x) = 1\}$. Generally, it is more useful to consider the the \textbf{$\mathcal{H}\Delta\mathcal{H}$-divergence}, specifically, where Ben-David et al. \cite{ben2010theory} define the symmetric difference hypothesis class $\mathcal{H}\Delta\mathcal{H}$ as the set of functions characteristic to disagreements between hypotheses.\footnote{Specifically, $g \in \mathcal{H} \Delta \mathcal{H} \quad \Leftrightarrow \quad g(x) = h_1(x) \oplus h_2(x) = \lvert h_1(x) - h_2(x)\rvert \quad\textrm{for}\quad h_1, h_2 \in \mathcal{H}$} This special case of the $\mathcal{H}$-divergence will be the measure of divergence in all considered bounds.

\subsubsection{The Motivating Bound}
We can now present the result of Ben-David et al. \cite{ben2010theory} based on the triangle inequality of classification error \cite{crammer2007learning, ben2007analysis}. 
This bound is the key motivation behind DANN \cite{ganin2015unsupervised}. For proof and a discussion on sample complexity, see Appendix~\ref{sec:appendix_a}.
\begin{theorem}[modified from Ben-David et al. \cite{ben2010theory}; Theorem 2]
\label{thm:bendavid}
Let $\mathcal{X}$ be a space and $\mathcal{H}$ be a class of hypotheses corresponding to this space. Suppose $\mathbb{P}$ and $\mathbb{Q}$ are distributions over $\mathcal{X}$. Then for any $h \in \mathcal{H}$,
\begin{equation}
    \mathcal{E}_\mathbb{Q}(h) \leq \lambda + 
    \mathcal{E}_\mathbb{P}(h) +
    \tfrac{1}{2} d_{\mathcal{H} \Delta \mathcal{H}}(\mathbb{Q}, \mathbb{P})
\end{equation}
with $\lambda$ the error of an ideal joint hypothesis for $\mathbb{Q}$, $\mathbb{P}$.
\end{theorem}
This statement provides an upper bound on the target-error. Thus, minimizing this upper bound is a good proxy for the minimization of the target-error itself. The first term $\lambda$ is a property of the dataset and hypothesis class which we typically assume to be small, but should not be ignored. As Ben-David et al. \cite{ben2010theory} note, this may be interpreted as a realizability assumption which requires the existence of some hypothesis in our search space that does well on both distributions (simultaneously). If this hypothesis does not exist, we cannot hope to do adaptation by minimizing the source-risk \cite{david10a}. Notably, $\lambda$ also plays an important role in algorithms like DANN which modify the distributions over which they learn since these algorithms implicitly change $\lambda$. We discuss this issue in detail in Section~\ref{sec:dann_limitations}. 

The latter terms are more explicitly controllable. The source-error $\mathcal{E}_\mathbb{P}(h)$ can be minimized as usual by Empirical Risk Minimization (ERM). The divergence can be empirically computed using another result of Ben-David et al. \cite{ben2010theory}. While we give this result in the Appendix (Proposition~\ref{prop:hdivemp} and Proposition~\ref{prop:generror}, respectively), previous interpretation by Ganing and Lempitsky~\cite{ganin2015unsupervised} suggests to minimize the divergence by learning indiscernible representations of the distributions -- i.e., aligning the domains.\footnote{Note, the motivation of this representation learning is not entirely precise. In fact, this is the cause of the issues we discuss later in Section~\ref{sec:dann_limitations}. } As we describe in the following, this may be accomplished by maximizing the errors of a domain discriminator trained to distinguish the distributions.

\subsubsection{The DANN Algorithm}
Ganin et al. \cite{ganin2015unsupervised} separate the neural network used to learn the task into a \textit{feature extractor} network $r_\theta$ and \textit{task-specific} network $c_\sigma$, parameterized respectively by $\theta$ and $\sigma$. A binary domain discriminator $d_\mu$ outputting probabilities is trained to distinguish between the source and target distribution based on the representation learned by $r_\theta$. Meanwhile, $r_\theta$ is trained to learn a representation that is not only useful for the task at hand, but also adept at ``\textit{fooling}'' the domain discriminator (i.e., maximizing its errors). In details, given an empirical sample $\hat{\mathbb{P}} = (x_i)_{i=1}^n$ from the source distribution $\mathbb{P}$ and a sample $\hat{\mathbb{Q}} = (x'_i)_{i=1}^n$ from the target distribution $\mathbb{Q}$, the domain adversarial training objective is described 
\begin{equation}
\begin{split}
    \min_\mu \max_\theta  \ 
    \frac{1}{2n} \sum_{i=1}^n
    \mathcal{L}_D(\mu, \theta, x_i, 0) + \mathcal{L}_D(\mu, \theta, x'_i, 1)
\end{split}
\end{equation}
where
\begin{equation}
\begin{split}
     -\mathcal{L}_D(\mu, \theta, x, y) = &(1-y) \log (1 - d_\mu \circ r_\theta (x))
    + y \log (d_\mu \circ r_\theta (x)).
\end{split}
\end{equation}
By this specification, $d_\mu \circ r_\theta (x)$ is meant to estimate the probability $x$ was drawn from $\mathbb{Q}$ and $\mathcal{L}_D$ represents the binary cross-entropy loss for a domain discriminator trained to distinguish $\mathbb{P}$ and $\mathbb{Q}$. Combining this with a task-specific loss $\mathcal{L}_T^\mathbb{P}$ we get the formulation given by Ganin et al. \cite{ganin2015unsupervised}
\begin{equation}
\label{eqn:dann_obj}
\begin{split}
    \min_{\sigma, \theta} & \max_\mu  \frac{1}{2n} \sum_{i=1}^n \mathcal{L}_T^\mathbb{P}(\sigma, \theta, x_i) - \frac{\lambda}{2n} \sum_{j=1}^n \mathcal{L}_D(\mu,\theta, x_j, 0) + \mathcal{L}_D(\mu, \theta, x'_j, 1)
\end{split}
\end{equation}
where $\lambda$ (in this context) is a trade-off parameter. The above is generally implemented by simultaneous gradient descent. We remark a solution to this optimization problem is easily approximated by incorporating a Gradient Reversal Layer (GRL) between $r_\theta$ and $d_\mu$ \cite{ganin2015unsupervised}.
\subsection{In Domain Generalization}
Recent adaptions to the above formulation have been proposed in context of DG. Here, we focus on the proposal of Matsuura and Harada \cite{matsuura2019domain} since their empirical results are one of the more competitive DG methods to date. In DG, since \textit{no} access to $\mathbb{Q}$ is given, one cannot actually compute $\mathcal{L}_D$ as described above -- it assumes at least unlabeled examples from $\mathbb{Q}$. Given this, Matsuura and Harada \cite{matsuura2019domain} propose a modification which operates on $k$ source samples
\begin{equation}
\label{eqn:lsd}
  -\mathcal{L}_{SD}(\mu, \theta, x, y) = \sum_{i=1}^k 1[i = y] \log ((d_\mu \circ r_\theta (x))_i)
\end{equation}
where $1[\cdot]$ is the indicator function. Now, $d_\mu$ is a multi-class domain discriminator trained to distinguish between \textit{sources}; it outputs the estimated probabilities that $x$ is drawn from each source. Hence, $\mathcal{L}_{SD}$ is essentially a multi-class cross-entropy loss. Given the source samples $\hat{\mathbb{P}}_j = (x_i^j)_{i=1}^n \ \forall j \in [k]$ drawn respectively from the source distributions $\mathbb{P}_1, \mathbb{P}_2, \ldots, \mathbb{P}_k$, we substitute this into Eq.~\eqref{eqn:dann_obj}:
\begin{equation}
\label{eqn:sdann_obj}
\begin{split}
 \min_{\sigma, \theta} \max_\mu \ \frac{1}{kn} \sum_{i=1}^n \sum_{j=1}^k \mathcal{L}_T^{\mathbb{P}_j}(\sigma, \theta, x_i^j)  + \frac{\lambda}{kn} \sum_{i=1}^n \sum_{j=1}^k \mathcal{L}_{SD}(\mu, \theta, x_i^j, j)
\end{split}
\end{equation}
which gives a domain adversarial training objective aimed at aligning the sources (while also maintaining good task performance). Hereon, we often refer to this as a \textit{source-source} DANN, rather than a \textit{source-target} DANN as was given in Eq.~\eqref{eqn:dann_obj}. On the surface, there seems to be no justification for the source-source DANN. If we recall the interpretation of Theorem~\ref{thm:bendavid}, there is one key difference: rather than aligning the source and target domains $\mathbb{P}$ and $\mathbb{Q}$ as suggested by the divergence term in Theorem~\ref{thm:bendavid}, the objective in Eq.~\eqref{eqn:sdann_obj} aligns source domains $\mathbb{P}_i$ and $\mathbb{P}_j \ \forall (i,j) \in [k]^2$ whose divergences do \textit{not} appear in the upper bound. Thus, the motivating argument is lost in this new formulation. If we look to recent literature, preliminary theoretical work to motivate this modification of DANN does exist \cite{albuquerque2019adversarial}. We start from this work in the derivation of our own results.

\subsection{A Gap Between Theory and Algorithm}
\label{sec:dann_limitations}
To be totally precise, the algorithm given above does not actually minimize $d_{\mathcal{H} \Delta \mathcal{H}}(\mathbb{P}_i, \mathbb{P}_j)$ for any $i,j$. As we have noted, the idea to ``align domains'' through a common feature representation is simply an interpretation following the convention of Ganin and Lempitsky \cite{ganin2015unsupervised}. If the class from which we select $d_\mu$ is $\mathcal{G}$ and the class from which we select $r_\theta$ is $\mathcal{F}$, the algorithm actually approximates minimization of $d_{\mathcal{G} \Delta \mathcal{G}}(\mathbb{P}_i \circ r_\theta^{-1}, \mathbb{P}_j \circ r_\theta^{-1})$ with respect to $\theta$. Here, the notation $\mathbb{P}_i \circ r_\theta^{-1}$ denotes the pushforward of $\mathbb{P}_i$ by $r_\theta$ which is  (intuitively) the image of $\mathbb{P}_i$ in the feature space. While this technicality will be unimportant for our discussions in the remainder of this text, it can potentially have significant negative ramifications. So, we discuss it in some detail here.

In particular, this gap between theory and algorithm implies that learning indiscernible representations of the source and target distributions while also minimizing the source error is not always sufficient for reducing the bound in Theorem~\ref{thm:bendavid}. The problem arises because the ideal joint error (which is usually assumed small in the original problem) does not always \textit{remain} small after feature transformation as in DANN. That is, while the ideal-joint error between $\mathbb{P}_i$ and $\mathbb{P}_j$ may be small, this may not be true of $\mathbb{P}_i \circ r_\theta^{-1}$ and $\mathbb{P}_j \circ r_\theta^{-1}$. This fact was recently observed independently by Johansson et al. \cite{johansson2019support} and Zhao et al. \cite{zhao2019learning}. Johansson et al. point out that learning a particular feature representation will always increase the ideal joint error (as compared to the original problem) whenever this feature representation is not invertible. Zhao et al. compliment this result by providing a lowerbound on target error in case the marginal label distributions\footnote{The marginal label distribution of the source or target is, formally, the pushforward of the source or target distribution by the respective labeling function.} have large deviation. In particular, the Jenson-Shannon (JS) divergence between the the label distributions should be at least as large as the JS divergence between the source and target feature distributions for the lowerbound to hold. If it is, the lowerbound shows simultaneous minimization of the source-error and the $\mathcal{H}\Delta\mathcal{H}$-divergence actually increases target-error.

In practice, as we are aware, it is not clear to what extent non-invertible feature representations increase the ideal joint error. Further, it is not easy to test whether the JS-divergence of the label distributions is larger than the JS-divergence of the source and target feature distributions. For this reason, in this work, we will simply assume the ideal joint error remains small after feature transformation; i.e., we do not explicitly consider any settings in which there are negative ramifications of the known gap between theory and algorithm for DANN. If these issues are of significant concern for a particular application (i.e., if the marginal label shift is known to be large), a recent modification of DANN which uses importance weighting has been proposed by Tachet des Combes, Zhao, et al. \cite{combes_and_zhao}. This modification aims to correct the short-comings of standard DANN in case of label-shift. While we do not explicitly experiment with this method, our theoretical discussion and algorithmic extension still apply in context of this variation on DANN.

%% file: 3_ours_theory.tex
Our discussion of source-source DANN for DG begins with the motivating target-error bound proposed by Albuquerque et al.~\cite{albuquerque2019adversarial}. Originally, given a set of source distributions $\{\mathbb{P}_i\}$, the bound uses the set of mixture distributions having these sources as components -- we refer to this set as $\mathcal{M}$. Below, we consider a more general adaptation of this result. Although the proof strategy is largely similar, we do provide proof for this more general re-statement. 
\begin{proposition}{(adapted from Albuquerque et al. \cite{albuquerque2019adversarial}; Proposition 2)}
\label{prop:albeq_general}
Let $\mathcal{X}$ be a space and let $\mathcal{H}$ be a class of hypotheses corresponding to this space. Let $\mathbb{Q}$ and the collection $\{\mathbb{P}_i\}_{i=1}^k$ be distributions over $\mathcal{X}$ and let $\{\varphi_i\}_{i=1}^k$ be a collection of non-negative coefficients with $\sum_i \varphi_i = 1$. Let the object $\mathcal{O}$ be a set of distributions such that for every $\mathbb{S} \in \mathcal{O}$ the following holds
\begin{equation}
\label{eqn:the_main_condition}
\sum\nolimits_i \varphi_i d_{\mathcal{H}\Delta\mathcal{H}}(\mathbb{P}_i, \mathbb{S}) \leq \max\nolimits_{i,j} d_{\mathcal{H}\Delta \mathcal{H}}(\mathbb{P}_i, \mathbb{P}_j). 
\end{equation}
Then, for any $h \in \mathcal{H}$,
\begin{equation}
\label{eqn:albeq_general}
\begin{split}
\mathcal{E}_\mathbb{Q}(h) \leq \lambda_\varphi +  \sum\nolimits_i \varphi_i  \mathcal{E}_{\mathbb{P}_i}(h)  & + \tfrac{1}{2}\min\nolimits_{\mathbb{S} \in \mathcal{O}}d_{\mathcal{H}\Delta\mathcal{H}}(\mathbb{S}, \mathbb{Q})
\\ & + \tfrac{1}{2}\max\nolimits_{i,j}  d_{\mathcal{H}\Delta \mathcal{H}}(\mathbb{P}_i, \mathbb{P}_j)
\end{split}
\end{equation}
where $\lambda_\varphi = \sum_i \varphi_i \lambda_i$ and each $\lambda_i$ is the error of an ideal joint hypothesis for $\mathbb{Q}$ and $\mathbb{P}_i$.
\end{proposition}
\begin{proof}
Let $h \in \mathcal{H}$. For each $\mathbb{P}_i$ apply Theorem \ref{thm:bendavid} and multiply the equation by $\varphi_i$ to achieve
\begin{equation}
\varphi_i\mathcal{E}_\mathbb{Q}(h) \leq \varphi_i \lambda_i +
    \varphi_i\mathcal{E}_{{\mathbb{P}}_i}(h) +
    \frac{\varphi_i}{2}  d_{\mathcal{H} \Delta \mathcal{H}}(\mathbb{Q}, \mathbb{P}_i)
\end{equation}
Taking $\lambda_\varphi = \sum_i \varphi_i \lambda_i$, we may sum over all $k$ of these inequalities as below 
\begin{equation}
\sum\nolimits_i \varphi_i\mathcal{E}_\mathbb{Q}(h) \leq  \lambda_\varphi + \sum\nolimits_i
    \varphi_i\mathcal{E}_{{\mathbb{P}}_i}(h)
    +
    \frac{\varphi_i}{2}  d_{\mathcal{H} \Delta \mathcal{H}}(\mathbb{Q}, \mathbb{P}_i). 
\end{equation}
Since $\sum_i \varphi_i = 1$ we can rewrite this as 
\begin{equation}
\mathcal{E}_\mathbb{Q}(h) \leq  \lambda_\varphi + \sum\nolimits_i \varphi_i\mathcal{E}_{{\mathbb{P}}_i}(h) + \frac{1}{2}\sum\nolimits_i \varphi_i d_{\mathcal{H} \Delta \mathcal{H}}(\mathbb{Q}, \mathbb{P}_i). 
\end{equation}
Now, for each $\mathbb{P}_i$, the following is true because the $\mathcal{H}$-divergence abides by the triangle inequality
\begin{equation}
    d_{\mathcal{H} \Delta \mathcal{H}}(\mathbb{Q}, \mathbb{P}_i) \leq d_{\mathcal{H} \Delta \mathcal{H}}(\mathbb{Q}, \mathbb{S}^*) + d_{\mathcal{H} \Delta \mathcal{H}}(\mathbb{S}^*, \mathbb{P}_i) 
\end{equation}
where 
\begin{equation}
\mathbb{S}^* \in \argmin\nolimits_{\mathbb{S}\in\mathcal{O}} d_{\mathcal{H} \Delta \mathcal{H}}(\mathbb{Q}, \mathbb{S}).
\end{equation}
Since this is true for each $\mathbb{P}_i$, we may write
\begin{equation}
\begin{split}
\label{eqn:extra_eq_to_explain}
\frac{1}{2}\sum\nolimits_i \varphi_i d_{\mathcal{H} \Delta \mathcal{H}}(\mathbb{Q}, \mathbb{P}_i) & \leq \frac{1}{2}\sum\nolimits_i \varphi_i d_{\mathcal{H} \Delta \mathcal{H}}(\mathbb{Q}, \mathbb{S}^*) + \frac{1}{2}\sum\nolimits_i \varphi_i d_{\mathcal{H} \Delta \mathcal{H}}(\mathbb{S}^*, \mathbb{P}_i) \\
& = \frac{1}{2} d_{\mathcal{H} \Delta \mathcal{H}}(\mathbb{Q}, \mathbb{S}^*) + \frac{1}{2}\sum\nolimits_i \varphi_i d_{\mathcal{H} \Delta \mathcal{H}}(\mathbb{S}^*, \mathbb{P}_i) \\
& \leq \frac{1}{2} d_{\mathcal{H} \Delta \mathcal{H}}(\mathbb{Q}, \mathbb{S}^*) + \frac{1}{2}\max\nolimits_{i,j} d_{\mathcal{H}\Delta \mathcal{H}}(\mathbb{P}_i, \mathbb{P}_j)
\end{split}
\end{equation}
where the last inequality is due to the choice $\mathbb{S}^* \in \mathcal{O}$. Recalling $\mathbb{S}^*$ is also a minimizer of $d_{\mathcal{H} \Delta \mathcal{H}}(\mathbb{Q}, \cdot)$ yields the result.
\end{proof}

As suggested by Albuquerque et al.~\cite{albuquerque2019adversarial}, interpreting this result provides a reasonable motivation for the use of source-source DANN in DG. The \textit{first term} is a convex combination of ideal-joint errors between each source and the target. As before, we assume this is small and remains small after feature transformation by $r_\theta$ when we apply DANN; i.e., recall Section~\ref{sec:dann_limitations}. Later, we discuss some differences between the ideal-error terms we give in our bound and the ideal-error terms in the original bound of Albuquerque et al.~\cite{albuquerque2019adversarial}. The \textit{second term} is a convex combination of the source errors. ERM on a mixture of the sources is appropriate for controlling this term. In both of the previous convex sums, the coefficients are assumed to be fixed, but arbitrary, replicating a natural data generation process where amounts of data from each source are not assumed. Ben-David et al.~\cite{ben2010theory} model data arising from multiple sources in this way and provide generalization bounds as well. For the \textit{third term}, when $\mathcal{O}$ is fixed as the set of mixtures $\mathcal{M}$, Albuquerque et al.~\cite{albuquerque2019adversarial} suggest this term demonstrates the importance of diverse source distributions, so that the unseen target $\mathbb{Q}$ might be ``near" $\mathcal{M}$. We extend this discussion later, showing how this term can change \textit{dynamically} throughout the training process. \textit{The final term} is a maximum over the source-source divergences. Application of the interpretation by Ganin and Lempitsky~\cite{ganin2015unsupervised} -- to align domains through representation learning -- motivates the suggestion of Matsuura and Harada~\cite{matsuura2019domain} to maximize the errors of a multi-class (source-source) domain discriminator. A more precise application might be to train all combinations of binary domain discriminator, but as Albuquerque et al.~\cite{albuquerque2019adversarial} point out, this leads to a polynomial number of discriminators. As a practical surrogate, we opt to employ the best empirical strategy to date \cite{matsuura2019domain}. Another option might be to instead use a collection of one-versus-all classifiers in place of a multi-class classifier \cite{albuquerque2019adversarial}. Note, neither method \textit{precisely} minimizes Eq.~\eqref{eqn:albeq_general}, so we treat this as an implementation choice.

\paragraph{A Remark on Differences} As mentioned briefly, a reader familiar with the original statement of Albuquerque et al.~\cite{albuquerque2019adversarial} will notice two differences: 1) rather than limiting consideration to the set of mixtures $\mathcal{M}$, this statement holds for all sets $\mathcal{O}$ which satisfy Condition~\eqref{eqn:the_main_condition} and 2) $\lambda_\varphi$ is a different quantity for the ideal joint-error between $\mathbb{Q}$ and $\{\mathbb{P}_i\}$. 

On the latter point, rather than $\lambda_\varphi$, Albuquerque et al.~\cite{albuquerque2019adversarial} use the following definition of the ideal joint error given by Zhao et al.~\cite{zhao2018adversarial} as below
\begin{equation}
    \lambda_* = \min_{h \in \mathcal{H}} \mathcal{E}_\mathbb{Q}(h) + \mathcal{E}_{\mathbb{S}^*}(h)
\end{equation}
where $\mathbb{S}^* \in \mathcal{M}$ is the mixture distribution closest to $\mathbb{Q}$. As the original statement of Albuquerque et al.~\cite{albuquerque2019adversarial} defines $\mathcal{O} = \mathcal{M}$, this definition is a perfectly reasonable choice. But, since our re-statement considers more general objects $\mathcal{O}$, we have removed this dependence on $\mathcal{M}$. As is visible in the proof, $\lambda_\varphi$ does remove this dependence. In general, $\lambda_*$ and $\lambda_\varphi$ are incomparable. If one attempts to compare them, it will become evident that some assumptions must be made -- e.g., on the relationship between the $\{\varphi_i\}_i$ (which are arbitrary but fixed) and the coefficients used to form the mixture for $\mathbb{S}^*$ (which are dependent on $\mathbb{Q}$). One reason to prefer $\lambda_\varphi$ is that it does not require a single hypothesis to have low error on \textit{all} sources simultaneously. Ben-David et al.~\cite{ben2010theory} provide a larger discussion on the benefits of various approaches when combining data from multiple sources.

The former difference is of primary interest in this paper. Condition~\eqref{eqn:the_main_condition} may be considered to be the \textit{key} fact about $\mathcal{M}$ which allows the derivation of Eq.~\eqref{eqn:albeq_general}. By identifying this, we open the possibility of considering more general objects satisfying Condition~\eqref{eqn:the_main_condition}. In the following, we demonstrate the existence of such objects $\mathcal{O}$ and discuss the benefit they add.

\subsection{Beyond Mixture Distributions}
\begin{figure}
    \centering
    \includegraphics[width=.4\linewidth]{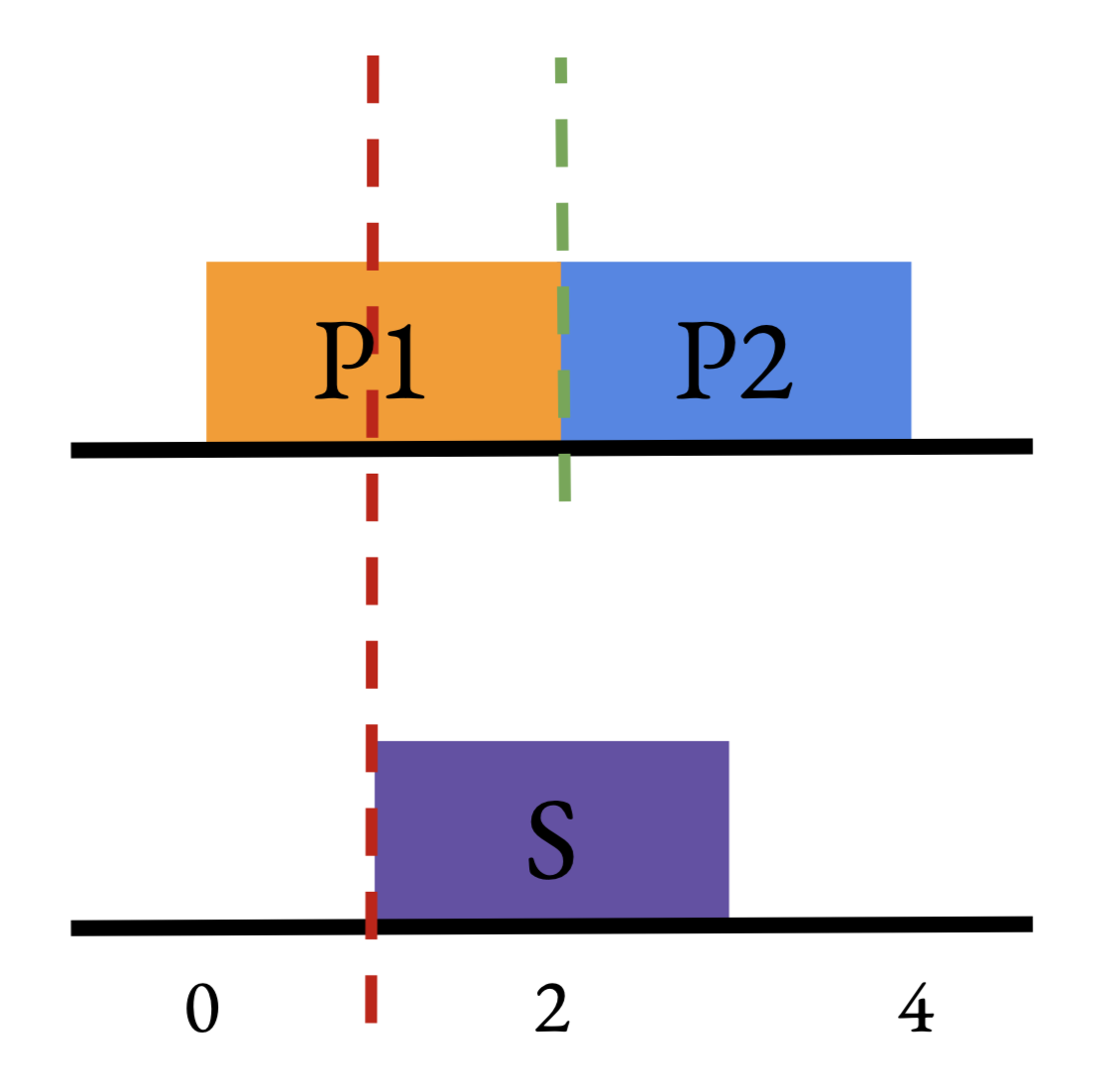}
    \caption{A visualization of Example~\ref{eg:counter_eg}. Best viewed in color. The green line gives the value $b$ of a hypothesis in $\{h_{a,b}(.)\}_{(a,b)}$ with $a \leq 0$. Such a hypothesis would perfectly discern $\mathbb{P}_1$ and $\mathbb{P}_2$. From this, it follows that $d_{\mathcal{H}\Delta \mathcal{H}}(\mathbb{P}_1, \mathbb{P}_2)=2$ because a hypothesis in $\{h_{a,b}(.)\}_{(a,b)}$ can achieve 2 and 2 is the maximum value for any divergence. Note, from this, it already follows that Eq~\eqref{eqn:eg_eq} holds because each term on the right-hand-side is bounded above by 2, and therefore, so is their convex combination. Still, we can analyze the example further. If we imagine the red line also gives the value $b$ of a hypothesis in $\{h_{a,b}(.)\}_{(a,b)}$ with $a \leq 0$ and slide it back and forth, we can never perfectly discern $\mathbb{P}_1$ or $\mathbb{P}_2$ from $\mathbb{S}$ and therefore we will never achieve the maximum divergence 2.}
    \label{fig:example}
\end{figure}
Consideration of general objects $\mathcal{O}$ which satisfy Condition~\eqref{eqn:the_main_condition} is only useful if such objects exist (besides $\mathcal{M}$). The following example provides proof. See Figure~\ref{fig:example} for an illustrative picture.

\begin{example} 
\label{eg:counter_eg}
Let $\mathcal{X}$ be the real line $(-\infty, \infty)$ and let $\mathcal{H}$ be the set of hypotheses $\{h_a(.)\}_{a \in \mathbb{R}}$ where $h_a(.)$ is characteristic to the ray $(-\infty, a]$. 
Then, $\mathcal{H}\Delta\mathcal{H}$ is the set of hypotheses $\{h_{a,b}(.)\}_{(a,b) \in \mathbb{R}^2}$ where $h_{a,b}(.)$ is characteristic to the interval $[a,b]$.
Let $\mathbb{P}_1$ be the uniform distribution $\mathcal{U}(0,2)$, let $\mathbb{P}_2$ be  $\mathcal{U}(2,4)$, and let $\mathbb{S}$ be  $\mathcal{U}(1,3)$. Then $\mathbb{S}$ is not a mixture distribution of the components $\mathbb{P}_1$ and $\mathbb{P}_2$, but 
\begin{equation}
\begin{split}
\label{eqn:eg_eq}
2 & =  \max\nolimits_{i,j} d_{\mathcal{H}\Delta \mathcal{H}}(\mathbb{P}_i, \mathbb{P}_j) \geq \sum\nolimits_i \varphi_i d_{\mathcal{H}\Delta \mathcal{H}}(\mathbb{P}_i, \mathbb{S})
\end{split}
\end{equation}
for all non-negative coefficients $\{\varphi_i\}_i$ which sum to 1. 
\end{example}
In the context of this example, we might consider the object $\mathcal{O} = \mathcal{M} + \{\mathbb{S}\}$ to quickly see that more than just $\mathcal{M}$ can satisfy Condition~\eqref{eqn:the_main_condition}. If $\mathbb{S}$ is a unique minimizer of the third term in Eq.~\eqref{eqn:albeq_general} and does not increase the final term, then using $\mathcal{O}$ in place of $\mathcal{M}$ actually produces a strictly tighter bound. Later we more generally expand on this and other benefits of considering $\mathcal{O} \neq \mathcal{M}$.

\begin{figure}
    \centering
    \includegraphics[width=.4\linewidth]{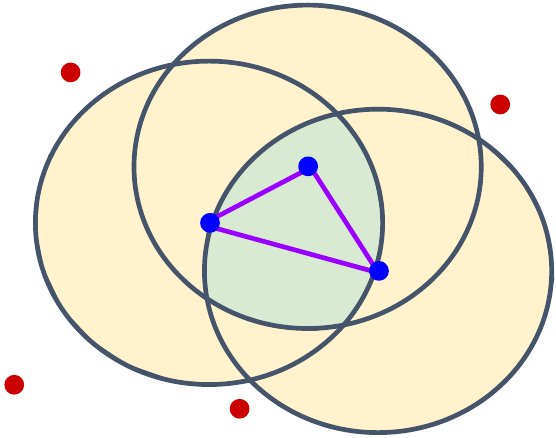}
    \caption{An informal visualization. Blue dots represent sources. Purple lines define the boundaries of $\mathcal{M}$. Grey lines give the boundaries of the closed $\mathcal{H},\rho$-balls around each source (defined in Proposition~\ref{thm:balls_in_distr_space}). Green colored areas define the boundary of $\bigcap\nolimits_i \mathcal{B}_\rho(\mathbb{P}_i)$. Distributions within the yellow area \textit{may} satisfy Condition~(\ref{eqn:the_main_condition}). Distributions outside the yellow area (red dots) do not satisfy Condition~(\ref{eqn:the_main_condition}).}
    \label{fig:viz}
\end{figure}

Still, one simple example cannot fully justify the existence of useful $\mathcal{O} \neq \mathcal{M}$. For a more general perspective, it is useful to think of things geometrically. Albuquerque et al.~\cite{albuquerque2019adversarial} often refer to $\mathcal{M}$ as the \textit{convex-hull} of the sources. In this same vein, we point out that $d_{\mathcal{H}\Delta\mathcal{H}}$ is a \textit{pseudometric}\footnote{In Appendix~\ref{sec:appendix_a}, we show the \textit{commonly used} fact that $d_{\mathcal{H}\Delta\mathcal{H}}$ possesses a triangle-inequality. Symmetry and evaluation to $0$ for identical distributions are easy to see.} and therefore, shares most of the nice properties required of \textit{metrics} used in the vast mathematical literature on metric spaces. Viewing a metric space as a \textit{topological space}, it is common to think of open balls as the ``the fundamental unit" or ``basis" of the metric space. Loosely, borrowing this idea, we can define the (closed) $\mathcal{H},\rho$-ball as below
\begin{equation}
   \mathcal{B}_\rho(\mathbb{P}) = \{\mathbb{S} \mid  d_{\mathcal{H}\Delta\mathcal{H}}(\mathbb{P}, \mathbb{S}) \leq \rho \}.
\end{equation}
Using this object, the following result provides some useful information on the types of objects $\mathcal{O}$ which satisfy Condition~\eqref{eqn:the_main_condition}. See Figure~\ref{fig:viz} for a helpful visualization of our results.

\begin{proposition}
\label{thm:balls_in_distr_space}
Let $\mathcal{X}$ be a space and let $\mathcal{H}$ be a class of hypotheses corresponding to this space. Let the collection $\{\mathbb{P}_i\}_{i=1}^k$ be distributions over $\mathcal{X}$ and let $\{\varphi_i\}_{i=1}^k$ be a collection of non-negative coefficients with $\sum_i \varphi_i = 1$. Now, set $\rho = \max\nolimits_{u,v} d_{\mathcal{H}\Delta \mathcal{H}}(\mathbb{P}_u, \mathbb{P}_v)$. We show three results,
\begin{enumerate}
\item $\mathcal{M} \subseteq \bigcap\nolimits_i \mathcal{B}_\rho(\mathbb{P}_i)$.
\item If $\mathbb{S} \in \bigcap\nolimits_i \mathcal{B}_\rho(\mathbb{P}_i)$, then Condition~\eqref{eqn:the_main_condition} holds.
\item If $\mathbb{S} \notin \bigcup\nolimits_i \mathcal{B}_\rho(\mathbb{P}_i)$, then Condition~\eqref{eqn:the_main_condition} fails to hold. 
\end{enumerate}
\end{proposition}

\begin{proof}
We begin with a proof of (1). Let $\mathbb{S} \in \mathcal{M}$ arbitrarily. The result follows by first observing, for all $\mathbb{P}_i$, 
\begin{equation}
d_{\mathcal{H}\Delta\mathcal{H}}(\mathbb{P}_i, \mathbb{S}) \leq \sum\nolimits_j \alpha_j d_{\mathcal{H}\Delta\mathcal{H}}(\mathbb{P}_i, \mathbb{P}_j) \leq \rho.
\end{equation}
The first inequality follows by a property of the $\mathcal{M}$ shown by  Albuquerque et al.~\cite{albuquerque2019adversarial}; for reference, we provide proof of this in Lemma~\ref{lem:mixtures_are_bounded_above_by_sums} in the Appendix. The second inequality follows because $\rho$ is defined as the largest source-source divergence. Now, if this is true for all $\mathbb{P}_i$, then $\mathbb{S}$ is by definition contained in every $\mathcal{H},\rho$-ball in the intersection $\bigcap\nolimits_i \mathcal{B}_\rho(\mathbb{P}_i)$. If an element is contained in every component set of an intersection, then it is contained in the intersection. And, we have shown (1).

Next, we show (2). By definition of $\mathcal{B}_\rho(\mathbb{P}_i)$, if $\mathbb{S} \in \mathcal{B}_\rho(\mathbb{P}_i)$ then $d_{\mathcal{H}\Delta\mathcal{H}}(\mathbb{P}_i, \mathbb{S}) \leq \rho$. Since $\mathbb{S} \in \bigcap\nolimits_i \mathcal{B}_\rho(\mathbb{P}_i)$ this is true for all $i \in [k]$. Then,
\begin{equation}
\begin{split}
\sum\nolimits_i \varphi_i d_{\mathcal{H}\Delta\mathcal{H}}(\mathbb{P}_i, \mathbb{S}) \leq \sum\nolimits_j \varphi_j \rho = \rho.
\end{split}
\end{equation}
We again recall that $\rho = \max\nolimits_{u,v} d_{\mathcal{H}\Delta \mathcal{H}}(\mathbb{P}_u, \mathbb{P}_v)$. Hence, we have shown (2).

Finally, we demonstrate (3). To see this, note that if $\mathbb{S} \notin \bigcup\nolimits_i \mathcal{B}_\rho(\mathbb{P}_i)$, then by definition for all $i$ we have that $d_{\mathcal{H}\Delta\mathcal{H}}(\mathbb{P}_i, \mathbb{S}) > \rho$. We follow the chain of inequalities below to arrive at our result
\begin{equation}
\begin{split}
& \sum\nolimits_i \varphi_i d_{\mathcal{H}\Delta\mathcal{H}}(\mathbb{P}_i, \mathbb{S}) \\
& > \sum\nolimits_i \varphi_i \rho \\
& = \max\nolimits_{i,j} d_{\mathcal{H}\Delta\mathcal{H}}(\mathbb{P}_i, \mathbb{P}_j).
\end{split}
\end{equation}
Hence, we have shown (3) and are done. 
\end{proof}

\textit{Statements 1 and 2} in conjunction show there are intuitive objects $\mathcal{O}$ -- i.e., $\bigcap\nolimits_i \mathcal{B}_\rho(\mathbb{P}_i)$ -- which both contain $\mathcal{M}$ and satisfy Condition~\eqref{eqn:the_main_condition}. \textit{Statement 3} provides an intuitive boundary for $\mathcal{O}$. Thus, comparison of $\mathcal{O}$ to the union and intersection of closed balls, respectively, provides necessary and sufficient conditions for satisfying Condition~\eqref{eqn:the_main_condition}.

\subsection{The Benefits of Looking Beyond Mixtures}
While the above discussion is useful in its own right, a more careful discussion of practical ramifications is needed. 

\paragraph{Computationally Tighter Bounds}
First, we point out that different objects $\mathcal{O}$ can lead to computationally tighter bounds in Eq.~\eqref{eqn:albeq_general}. For a concrete example, we prove $\bigcap\nolimits_i \mathcal{B}_\rho(\mathbb{P}_i)$ can lead to tighter bounds than $\mathcal{M}$ below. The proof follows a similar logic as presented following Example~\ref{eg:counter_eg}. In fact, for Example~\ref{eg:counter_eg}, it is true that $\bigcap\nolimits_i \mathcal{B}_\rho(\mathbb{P}_i)$ contains $\mathcal{M} + \{\mathbb{S}\}$, and thus, may reap the discussed benefit.
\begin{proposition}
\label{prop:ours_is_tighter}
Let $\mathcal{X}$ be a space and let $\mathcal{H}$ be a class of hypotheses corresponding to this space. Let $\mathbb{Q}$ and the collection $\{\mathbb{P}_i\}_{i=1}^k$ be distributions over $\mathcal{X}$. Let $\mathbb{P}^*$ be the distribution in $\bigcap\nolimits_i \mathcal{B}_\rho(\mathbb{P}_i)$ closest to $\mathbb{Q}$ and let $\mathbb{S}^* \in \mathcal{M}$ be the mixture distribution closest to $\mathbb{Q}$. Then,
\begin{equation}
    d_{\mathcal{H}\Delta\mathcal{H}}(\mathbb{P}^*, \mathbb{Q}) \leq d_{\mathcal{H}\Delta\mathcal{H}}(\mathbb{S}^*, \mathbb{Q}).
\end{equation}
Now, further, suppose the only solution to   
\begin{equation}
\label{eqn:only_solution}
    \min_{\mathbb{P} \in \bigcap\nolimits_i \mathcal{B}_\rho(\mathbb{P}_i) }d_{\mathcal{H}\Delta\mathcal{H}}(\mathbb{P}, \mathbb{Q})
\end{equation}
is contained in $\bigcap\nolimits_i \mathcal{B}_\rho(\mathbb{P}_i) \setminus \mathcal{M}$. Then, we have
\begin{equation}
    d_{\mathcal{H}\Delta\mathcal{H}}(\mathbb{P}^*, \mathbb{Q}) <  d_{\mathcal{H}\Delta\mathcal{H}}(\mathbb{S}^*, \mathbb{Q}).
\end{equation}
\end{proposition}
\begin{proof}
To see the first claim, note by Proposition~\ref{thm:balls_in_distr_space}, $\mathcal{M} \subseteq \bigcap\nolimits_i \mathcal{B}_\rho(\mathbb{P}_i)$. So it is clear that
\begin{equation}
\label{eqn:checkpoint_2}
    \min_{\mathbb{P} \in \bigcap\nolimits_i \mathcal{B}_\rho(\mathbb{P}_i)} d_{\mathcal{H}\Delta\mathcal{H}}(\mathbb{P}, \mathbb{Q}) \leq  \min_{\mathbb{S} \in \mathcal{M}} d_{\mathcal{H}\Delta\mathcal{H}}(\mathbb{S}, \mathbb{Q}).
\end{equation}
Since $\mathbb{P}^*$ and $\mathbb{S}^*$ are arguments minimizing left- and right-hand-side, respectively, we are done. 

Now, we show the second claim. Equation~\ref{eqn:checkpoint_2} holds irregardless of our additional assumption, so we need only show that 
\begin{equation}
    \min_{\mathbb{P} \in \bigcap\nolimits_i \mathcal{B}_\rho(\mathbb{P}_i)} d_{\mathcal{H}\Delta\mathcal{H}}(\mathbb{P}, \mathbb{Q}) \neq \min_{\mathbb{S} \in \mathcal{M}} d_{\mathcal{H}\Delta\mathcal{H}}(\mathbb{S}, \mathbb{Q}).
\end{equation}
But this is clear because if we assume the contrary -- that the two quantities are equal -- the implication is that a solution to Equation~\ref{eqn:only_solution} is contained in $\mathcal{M}$, a contradiction. Therefore, we have our result.
\end{proof}

Now, for DANN, our hypothesis will usually be a neural network. In this case, the benefit of tightness may be considered irrelevant because the large VC-Dimension of neural networks \cite{bartlett2019nearly} is the dominant term in any bound on error (i.e., using the PAC framework). Still, this conversation is not complete without considering the recent success of PAC-Bayesian formulations (e.g., see Dziugaite et al.~\cite{dziugaite2017computing}) which provide much tighter bounds when the hypothesis is a stochastic neural network. In Appendix~\ref{sec:appendix_a}, we discuss a PAC-Bayesian distribution psuedometric \cite{germain2020pac} analogous to $d_{\mathcal{H} \Delta \mathcal{H}}$. Because this psuedometric shares the important properties of $d_{\mathcal{H} \Delta \mathcal{H}}$, these results are easily re-framed in this more modern formulation as well -- where tightness may be a primary concern. 

\paragraph{Intuitive Analysis}
Second, we point out that a particular object $\mathcal{O}$ can be easier to analyze. This fact will become evident as we develop an algorithmic extension to DANN for DG. Ultimately, we find that the novel object $\bigcap\nolimits_i \mathcal{B}_\rho(\mathbb{P}_i)$ may be manipulated to provide key motivating insights in algorithm design. 

\subsection{The $\mathcal{H}\Delta\mathcal{H}$-Divergence as a Dynamic Quantity}
\label{sec:dynamic}

As mentioned, Albuquerque et al.~\cite{albuquerque2019adversarial} interpret Proposition~\ref{prop:albeq_general} as showing the necessity of diverse source distributions to control the third term $\min_{\mathbb{S} \in \mathcal{O}} d_{\mathcal{H}\Delta\mathcal{H}}(\mathbb{S}, \mathbb{Q})$ when $\mathcal{O} = \mathcal{M}$. Logically, when distributions are heterogeneous, $\mathcal{M}$ presumably contains more elements, and so, the unseen target is more likely to be ``close." When $\mathcal{O} = \bigcap\nolimits_i \mathcal{B}_\rho(\mathbb{P}_i)$, this is easier to see because the size of $\mathcal{O}$ is directly dependent on the maximum divergence between the sources (by the definition of $\rho$). In particular, reducing the maximum divergence and re-computing $\mathcal{O}$ could lead to removal of a unique minimizer for $\min_{\mathbb{S} \in \mathcal{O}} d_{\mathcal{H}\Delta\mathcal{H}}(\mathbb{S}, \mathbb{Q})$.\footnote{Under conditions discussed later, the newly computed $\mathcal{O}$ will be a subset and this unique minimizer might be absent in this subset.} In the context of the DANN algorithm, this is worrisome. Namely, during training, the point of using DANN is to effectively reduce the maximum divergence between sources and we expect this divergence to be decreasing as the feature representations of the source distributions are modified. In fact, under mild assumptions, we can formally show that DANN acts like a contraction mapping, and therefore, can \textit{only} decrease the pairwise source-divergences. So, it is possible $\min_{\mathbb{S} \in \mathcal{O}} d_{\mathcal{H}\Delta\mathcal{H}}(\mathbb{S}, \mathbb{Q})$ increases as the changing object $\mathcal{O}$ shrinks during training. Below we consider gradient descent on a smooth proxy of the $\mathcal{H} \Delta \mathcal{H}$-Divergence in the simple, two-distribution case. The map $r_\theta$ acts as the feature extractor affected by DANN.
\begin{proposition}
\label{prop:contractions}
Let $\mathfrak{D}$ be a space of empirical samples over $\mathcal{X}$. Let $r_\theta : \mathcal{X} \to \mathcal{X}$ be a deterministic representation function parameterized by the real vector $\theta \in \mathbb{R}^m$. Further, denote by $r_\theta(\widehat{\mathbb{P}})$ the application of $r_\theta$ to every point of $\widehat{\mathbb{P}} \in \mathfrak{D}$. Fix $\widehat{\mathbb{P}}, \widehat{\mathbb{Q}} \in \mathfrak{D}$, let $\mathcal{L}: \mathfrak{D} \times \mathfrak{D} \to [0, \infty)$. Define $\ell(\theta) = \mathcal{L}(r_\theta(\widehat{\mathbb{P}}), r_\theta(\widehat{\mathbb{Q}}))$ and suppose it is differentiable with $K$-Lipschitz gradients. Further, suppose $\theta^*$ is the unique local minimum of $\ell$ on a bounded subset $\Omega \subset \mathbb{R}^m$. Then for $\theta \in \Omega$ such that $\theta \neq \theta^*$, the function $\tau: \Omega \to \mathbb{R}^m$ defined $\tau(\theta) = \theta - \gamma\nabla_\theta\ell(\theta)$ has the property
\begin{equation}
\label{eqn:elltaubound}
    \mathcal{L}(r_{\tau(\theta)}(\widehat{\mathbb{P}}), r_{\tau(\theta)}(\widehat{\mathbb{Q}})) \leq \beta_\theta \mathcal{L}(r_\theta(\widehat{\mathbb{P}}), r_\theta(\widehat{\mathbb{Q}}))
\end{equation}
for some constant $\beta_\theta$ dependent on $\theta$.
In particular, for all $\theta \in \Omega$, there is $\gamma$ so that 
$0 < \beta_\theta < 1$.
\end{proposition}
\begin{proof}
We proceed by first showing an import inequality for functions $\ell$ with the assumed properties, in particular, using a derivation presented by Wright~\cite{stephenwrightoptimization}. Note first, by Taylor's Theorem, for vectors $u,v \in \mathbb{R}^n$, we have 
\begin{equation}
    \begin{split}
        \ell(u + v) & = \ell(u) + \int_{0}^1 \nabla \ell(u + \xi v)^\mathrm{T}v \ d\xi \\
        & = \ell(u) + \nabla \ell(u)^\mathrm{T}v + \int_{0}^1 \nabla \left [\ell(u + \xi v) - \nabla \ell(u) \right ]^\mathrm{T}v \ d\xi \\
        & \leq \ell(u) + \nabla \ell(u)^\mathrm{T}v + \int_{0}^1 \lvert\lvert \nabla \ell(u + \xi v) - \nabla \ell(u) \rvert \rvert \ \lvert \lvert v \rvert \rvert \ d\xi \\
        & \leq \ell(u) + \nabla \ell(u)^\mathrm{T}v + \int_{0}^1  \xi K\lvert \lvert v \rvert \rvert^2 \ d\xi \\
        & = \ell(u) + \nabla \ell(u)^\mathrm{T}v + \frac{1}{2}K \lvert\lvert v \rvert \rvert^2.
    \end{split}
\end{equation}
where the first line, as mentioned, is by Taylor's Theorem, the second is by addition and subtraction of $\nabla \ell(u)^\mathrm{T}v$, the third is because the norm of a vector product is never larger than the vector product, and the fourth is by the Lipshitz property assumed on the gradients of $\ell$. 

With this inequality, we let $\theta \in \Omega$ with $\theta \neq \theta^*$. Taking $u = \theta$ and $v = -\gamma \nabla \ell(\theta)$ achieves
\begin{equation}
\label{eqn:elltauboundpf}
\begin{split}
    \ell(\tau(\theta)) & \leq \ell(\theta) - \gamma \nabla\ell(\theta)^\mathrm{T}\nabla\ell(\theta) + \frac{\gamma^2K}{2} \lvert\lvert \nabla\ell(\theta)\rvert\rvert^2 \\
    & = \ell(\theta) + \gamma (\tfrac{1}{2}\gamma K - 1)\lvert\lvert\nabla\ell(\theta)\rvert\rvert^2.
\end{split}
\end{equation}
Next, we note that for $\theta \neq \theta^*$ we have $0 \leq \ell(\theta^*) < \ell(\theta)$ because $\theta^*$ was assumed to be the \textit{unique} local minimum of $\Omega$. Then, we may set 
\begin{equation}
    \beta_\theta = 1 + \gamma \left (\tfrac{1}{2} \gamma K - 1\right) \frac{\lvert\lvert\nabla \ell(\theta)\rvert\rvert^2}{\ell(\theta)}
\end{equation}
which, in combination with Eq.~\eqref{eqn:elltauboundpf} yields our first desired result (Eq.~\eqref{eqn:elltaubound}).

Next, we show that for all $\theta \neq \theta^*$, we can pick $\gamma$ which forces $0 < \beta_\theta < 1$. We first note that it is sufficient to show
\begin{equation}
\label{eqn:gammasuff1}
    \frac{-\ell(\theta)}{\lvert\lvert\nabla \ell(\theta)\rvert\rvert^2} < \gamma \left (\tfrac{1}{2} \gamma K - 1\right) < 0
\end{equation}
since we may simply multiply by the reciprocal of the lower-bound and add one to realize the result. Next, we point out that there is some constant $M > 0$ such that $\vert\vert\nabla \ell(\theta)\vert\vert \leq KM$. This follows by
\begin{equation}
\label{eqn:gradbound}
    \lvert\lvert\nabla \ell (\theta)\rvert\rvert = \lvert\lvert\nabla \ell(\theta) - \nabla \ell (\theta^*)\rvert\vert \leq K \vert\vert\theta - \theta^*\vert\vert \leq KM 
\end{equation}
where the equality holds because $\theta^*$ is a local minimum, the first inequality holds by the assumed Lipshitz property, and the second inequality holds because $\Omega$ was assumed to be bounded. Without loss of generality, suppose $M \geq 1$ (Eq.~\eqref{eqn:gradbound} holds regardless). Then our problem reduces further. In particular, it suffices to pick $\gamma$ such that
\begin{equation}
\label{eqn:gammasuff2}
    \frac{-\ell(\theta)}{K^2M^2} < \gamma \left (\tfrac{1}{2} \gamma K - 1\right) < 0
\end{equation}
since this lower bound is larger than or equal to that of Eq.~\eqref{eqn:gammasuff1}. First, clearly, the upper bound holds when $0 < \gamma < \tfrac{2}{K}$, so this immediately restricts our choice of $\gamma$. For the lower bound, we consider two cases for the value of $\ell(\theta)$ and demonstrate there is $\gamma$ with $0 < \gamma < \tfrac{2}{K}$ in both.

First, suppose $\ell(\theta) \geq \tfrac{1}{2}KM^2$. Then, if $\tfrac{2}{K} > \gamma > \tfrac{1}{K}$ we have 
\begin{equation}
\begin{split}
    0 & > \gamma \left (\tfrac{1}{2} \gamma K - 1\right)
    > \frac{-1}{2K} = \frac{-KM^2}{2K^2M^2} > \frac{-\ell(\theta)}{K^2M^2}.
\end{split}
\end{equation}
Second, suppose $\ell(\theta) < \tfrac{1}{2}KM^2.$ Then if $\gamma$ is such that
\begin{equation}
    \frac{2}{K} > \gamma > \frac{1 - \sqrt{1 - \tfrac{2\ell(\theta)}{KM^2}}}{K}
\end{equation}
we have
\begin{equation}
\begin{split}
    & \gamma \left (\tfrac{1}{2} \gamma K - 1\right) + \frac{\ell(\theta)}{K^2M^2} \\
    & \qquad >  \frac{K}{2} \left( \frac{1 - \sqrt{1 - \tfrac{2\ell(\theta)}{KM^2}}}{K}\right)^2 - \frac{1 - \sqrt{1 - \tfrac{2\ell(\theta)}{KM^2}}}{K} + \frac{\ell(\theta)}{K^2M^2} \\
    & \qquad = 0 . 
\end{split}
\end{equation}
Subtracting $\tfrac{\ell(\theta)}{K^2M^2}$ from both sides of this inequality yields the desired lower bound. Further, we still have $\gamma < \frac{2}{K}$, so the desired upper bound holds and we have our result.

Then, in any case, for each $\theta \neq \theta^*$, we can select $\gamma$ so that $0 < \beta_\theta < 1$.
\end{proof}

A \textbf{key takeaway} from the above is the presence of competing objectives during training. These objectives require balance. While DANN reduces the source-divergences to account for the final term in Eq.~\eqref{eqn:albeq_general}, we should also (somehow) consider the diversity of our sources throughout training to account for the effected term $\min_{\mathbb{S} \in \mathcal{O}} d_{\mathcal{H}\Delta\mathcal{H}}(\mathbb{S}, \mathbb{Q})$. Another insight the reader gains (i.e., from reading the proof) is that the upper bound on $\gamma$ is constant and the lower bound goes to 0 as $\ell (\theta) \to 0$. An interpretation of these bounds suggests the practical importance of an annealing schedule on $\gamma$ during DANN training. In our own experiments, we anneal $\gamma$ by a constant factor (i.e., step decay).

%% file: 4_ours_algorithm.tex
Motivated by the argument presented in Section~\ref{sec:dynamic}, this section devises an extension to DANN.
While DANN acts to align domains, as noted, its success in the context of domain generalization is also dependent on the heterogeneity of the source distributions \textit{throughout} the training process. Therefore, in an attempt to balance these objectives, we propose an addition to source-source DANN which acts to diversify the sources throughout the training.
\paragraph{Theoretical Motivation} We recall the intersection of closed balls $\mathcal{O} = \bigcap\nolimits_i \mathcal{B}_{\rho} (\mathbb{P}_i)$; 
this is the main object of interest as it controls the size of the divergences in the upper bound of Proposition~\ref{prop:albeq_general}. More specifically, we are concerned with the quantity $\min_{\mathbb{P} \in  \bigcap\nolimits_i \mathcal{B}_\rho(\mathbb{P}_i)} d_{\mathcal{H}\Delta\mathcal{H}}(\mathbb{P}, \mathbb{Q})$.
Intuitively, if we want to reduce this quantity we should find some means to increase $\rho$. One might propose to accomplish this by modifying our source distributions -- e.g., through data augmentation --, but clearly, modifying our source distributions in an \textit{uncontrolled} manner is not wise. This ignores the structure of the space of distributions under consideration and whichever distribution governs our sampling from this space -- information that is, in part, given by our sample of sources itself. In this sense, while increasing $\rho$, we should preserve the structure of $\bigcap\nolimits_i \mathcal{B}_{\rho} (\mathbb{P}_i)$ as much as possible. Proposition~\ref{thm:augmentation_theory} identifies conditions we must satisfy if we wish to increase $\rho$ and modify our source distributions in a way that is \textit{guaranteed} to reduce the third term of the upperbound in Eq.~\eqref{eqn:albeq_general}. 
\begin{proposition}
\label{thm:augmentation_theory}
Let $\mathcal{X}$ be a space and let $\mathcal{H}$ be a class of hypotheses corresponding to this space. Let $\mathfrak{D}$ be the space of distributions over $\mathcal{X}$ and let the collection $\{\mathbb{P}_i\}_{i=1}^k$ and the collection $\{\mathbb{R}_i\}_{i=1}^k$ be contained in $\mathfrak{D}$. Now, consider the collection of mixture distributions $\{\mathbb{S}_i\}_i$ defined so that for each set $A$, $\mathrm{Pr}_{\mathbb{S}_i}(A) = \alpha \mathrm{Pr}_{\mathbb{P}_i}(A) + \beta \mathrm{Pr}_{\mathbb{R}_i}(A)$. Further, set $\rho = \max\nolimits_{i,j} d_{\mathcal{H}\Delta \mathcal{H}}(\mathbb{P}_i, \mathbb{P}_j)$ and  $\rho^* = \max\nolimits_{i,j} d_{\mathcal{H}\Delta \mathcal{H}}(\mathbb{S}_i, \mathbb{S}_j)$. Then 
$
\label{eqn:result_of_augmentation}
\bigcap\nolimits_i \mathcal{B}_\rho (\mathbb{P}_i) \subseteq \bigcap\nolimits_i \mathcal{B}_{\rho^*} (\mathbb{S}_i)
$
whenever
$
\rho^* - \beta \max_i d_{\mathcal{H}\Delta \mathcal{H}}(\mathbb{R}_i, \mathbb{P}_i) \geq \rho.$
\end{proposition}

\begin{proof}
Let $\mathbb{Q} \in \bigcap\nolimits_i \mathcal{B}_\rho (\mathbb{P}_i) $ be arbitrary. Then, by definition, for all $i$, we have that
\begin{equation}
d_{\mathcal{H}\Delta \mathcal{H}}(\mathbb{P}_i, \mathbb{Q}) \leq \rho. 
\end{equation}
Then,  for all $i$, we have 
\begin{equation}
\begin{split}
& d_{\mathcal{H}\Delta \mathcal{H}}(\mathbb{S}_i, \mathbb{Q})  \leq \alpha d_{\mathcal{H}\Delta \mathcal{H}}(\mathbb{P}_i, \mathbb{Q}) + \beta d_{\mathcal{H}\Delta \mathcal{H}}(\mathbb{R}_i, \mathbb{Q}) \\
& \leq \alpha \rho + \beta d_{\mathcal{H}\Delta \mathcal{H}}(\mathbb{R}_i, \mathbb{Q})  \\
& \leq \alpha \rho + \beta d_{\mathcal{H}\Delta \mathcal{H}}(\mathbb{R}_i, \mathbb{P}_i) + \beta d_{\mathcal{H}\Delta \mathcal{H}}(\mathbb{P}_i, \mathbb{Q})  \\
& \leq (\alpha + \beta) \rho +  \beta d_{\mathcal{H}\Delta \mathcal{H}}(\mathbb{R}_i, \mathbb{P}_i) \\
& = \rho +  \beta d_{\mathcal{H}\Delta \mathcal{H}}(\mathbb{R}_i, \mathbb{P}_i) \\
& \leq \rho +  \beta \max\nolimits_i d_{\mathcal{H}\Delta \mathcal{H}}(\mathbb{R}_i, \mathbb{P}_i) \\
& \leq \rho^*
\end{split}
\end{equation}
where the first inequality follows by Lemma~\ref{lem:mixtures_are_bounded_above_by_sums}, the second inequality follows because $\mathbb{Q} \in \bigcap\nolimits_i \mathcal{B}_\rho (\mathbb{P}_i)$ so the divergence is bounded by $\rho$ for all $i$, the third inequality follows because, in general, the $\mathcal{H}$-divergence abides by the triangle-inequality, the fourth inequality follows again because $\mathbb{Q} \in \bigcap\nolimits_i \mathcal{B}_\rho (\mathbb{P}_i)$, and the last inequality follows because we have assumed
\begin{equation}
\rho^* - \beta \max\nolimits_i d_{\mathcal{H}\Delta \mathcal{H}}(\mathbb{R}_i, \mathbb{P}_i) \geq \rho.
\end{equation}
Now, this is true for all $i$, so by definition of $\bigcap\nolimits_i \mathcal{B}_{\rho^*} (\mathbb{S}_i)$, we have that $\mathbb{Q} \in \bigcap\nolimits_i \mathcal{B}_{\rho^*} (\mathbb{S}_i)$. Since $\mathbb{Q}$ was an arbitrary element of $ \bigcap\nolimits_i \mathcal{B}_\rho (\mathbb{P}_i)$, we have shown  $\bigcap\nolimits_i \mathcal{B}_\rho (\mathbb{P}_i) \subseteq \bigcap\nolimits_i \mathcal{B}_{\rho^*} (\mathbb{S}_i)$ and we have our result.
\end{proof}

The above statement suggests that if we want to diversify our training distributions, we should train on a collection of modified source distributions $\{\mathbb{S}_i\}_i$. The modified distributions are mixture distributions whose components are pairs of our original source distributions $\{\mathbb{P}_i\}_i$ and new \textit{auxiliary} distributions $\{\mathbb{R}_i\}_i$. The choice of $\{\mathbb{R}_i\}_i$ is constrained to guarantee the new intersection $\bigcap\nolimits_i \mathcal{B}_{\rho^*} (\mathbb{S}_i)$ (with modified sources) contains the original intersection $\bigcap\nolimits_i \mathcal{B}_{\rho} (\mathbb{P}_i)$. Ultimately, this means we can guarantee $\min_{\mathbb{S} \in  \bigcap\nolimits_i \mathcal{B}_{\rho^*}(\mathbb{S}_i)} d_{\mathcal{H}\Delta\mathcal{H}}(\mathbb{S}, \mathbb{Q}) \leq \min_{\mathbb{P} \in  \bigcap\nolimits_i \mathcal{B}_\rho(\mathbb{P}_i)} d_{\mathcal{H}\Delta\mathcal{H}}(\mathbb{P}, \mathbb{Q})$. 

\paragraph{Algorithm} Empirically speaking, our modified source samples $\{\hat{\mathbb{S}}_i\}_i$ 
will be a mix of examples from the original sources $\{\mathbb{P}_i\}_i$ and the auxiliary distributions $\{\mathbb{R}_i\}_i$ -- drawn from each proportionally to the mixture weights $\alpha$ and $\beta$. We plan to \textit{generate} samples from the auxiliary distributions $\{\mathbb{R}_i\}_i$ and our interpretation of Proposition~\ref{thm:augmentation_theory} suggests we should do so subject to the constraint below 
\begin{equation}
\label{eqn:full_augmentation_objective}
\max\nolimits_{i,j} d_{\mathcal{H}\Delta \mathcal{H}}(\mathbb{S}_i, \mathbb{S}_j)- \beta \max\nolimits_i d_{\mathcal{H}\Delta \mathcal{H}}(\mathbb{R}_i, \mathbb{P}_i) \geq \rho.
\end{equation}
Because $\rho$ is a property of our original dataset, it is independent of the distributions $\{\mathbb{R}_i\}_i$. This suggests that we should generate $\{\hat{\mathbb{R}}_i\}_i$ to maximize the left hand side. Maximizing this requires: \textbf{(Req.I)}
maximizing the largest divergence between the new source samples $\{\hat{\mathbb{S}}_i\}_i$ and \textbf{(Req.II)} minimizing the largest divergence between our auxiliary samples $\{\hat{\mathbb{R}}_i\}_i$ and our original source samples  $\{\hat{\mathbb{P}}_i\}_i$. Algorithmically, we can coarsely approximate these divergences, again appealing to the interpretation provided by Ben-David et al.~\cite{ben2010theory} and Ganin and Lempitsky~\cite{ganin2015unsupervised}: \textbf{(Req.I)} requires that our domain discriminator make fewer errors when discriminating the new source samples $\{\hat{\mathbb{S}}_i\}_i$ and \textbf{(Req.II)} requires that the auxiliary samples $\{\hat{\mathbb{R}}_i\}_i$ and the original sources $\{\hat{\mathbb{P}}_i\}_i$ be indiscernible by our domain discriminator.

To implement these requirements, we modify our dataset through gradient descent. Suppose that $\hat{\mathbb{P}}_i$ is an empirical sample from the distribution $\mathbb{P}_i$. We can alter data-points $a^j \sim \hat{\mathbb{P}}_j$ to generate data-points $b^j \sim\hat{\mathbb{R}}_j$ by setting $x^j(0) = a^j$ and iterating the below update rule to minimize $\mathcal{L}_{SD}$ for $T$ steps
\begin{equation}
\label{eqn:update_rule_first_pass}
\begin{split}
x^j(t) \gets x^j(t-1) - \eta \nabla_{x} \mathcal{L}_{SD}(\mu, \theta, x^j(t-1), j)
\end{split}
\end{equation}
and then taking $b^j = x^j(T)$. \textit{Importantly}, we do not modify the domain labels in this modification. So, our updates satisfy requirement \textbf{(Req.I)} because minimization of $\mathcal{L}_{SD}$ approximates minimization of our domain discriminator's errors, and further, satisfy \textbf{(Req.II)} because $a_i$ and $b_i$ are identically labeled, so minimization of the domain discriminator's errors suggests that these examples should be indiscernible (i.e., assigned the same correct label). 

While this update rule seemingly accomplishes our algorithmic goals, we must recall the final upper bound we wish to minimize (see Eq.~\eqref{eqn:albeq_general}). The first two terms in this bound, $\lambda_\varphi$ and $\sum_i \varphi_i\mathcal{E}_{\mathbb{P}_i}(h)$,  relate to our \textit{classification} error -- i.e., to the task-specific network $c_\sigma$. If our generated distributions $\{\hat{\mathbb{R}}_i\}_i$ distort the underlying class information, these terms may grow uncontrollably. 
To account for this, we further modify the update rule of Eq.~\eqref{eqn:update_rule_first_pass} to minimize the change in the probability distribution output by the task classifier. We measure the change caused by our updates using the loss $\mathcal{L}_{KL}$ -- i.e., the KL-Divergence \citep{kullback1997information}. This gives the modified update rule  
\begin{equation}
\label{eqn:update_rule_second_pass}
\begin{split}
x^j_i(t) \gets & x^{j}_i(t-1) - \eta \nabla_{x}  \left[ \mathcal{L}_{SD}(\mu, \theta, x^{j}_i(t-1), j) \right. \\ & \quad\left. + \mathcal{L}_{KL}(c_\sigma \circ r_\theta (x^j_i(0)), c_\sigma \circ r_\theta (x^{j}_i(t-1)))\right].
\end{split}
\end{equation}

\begin{algorithm}
   \caption{DANNCE (DANN with Cooperative Examples)}
   \label{alg:example}
\begin{algorithmic}
   \State {\bfseries Input:} A collection of mini-batches $\{x_i^j\}_i$ with labels $\{y_i^j\}_i$ for each $j \in [k]$, classifier $c_\sigma$ parameterized by $\sigma$, feature extractor $r_\theta$ parameterized by $\theta$, domain discriminator $d_\mu$ parameterized by $\mu$
   \State \textbf{Parameters:} Perturbation probability $\beta$ between 0 and 1. Number of update steps $T$. Learning rate $\eta$.
   \State \textbf{Dependencies:} SourceSourceDANN (i.e., any DANN algorithm to optimize for main-text Eq.~8 given a batch)
   \For{$j=1:k$}
        \For{example $x_\ell^j \in$ batch $\{x_i^j\}_i$}
            \State $p \gets$ random uniform draw from $[0,1]$
            \If{$p \leq \beta$}
                \For{$t=1:T$}
                    \State Update $x^j_\ell(t)$ using Eq.~\eqref{eqn:update_rule_second_pass} \Comment{DANNCE Update}
                \EndFor
                \State $\{x_i^j\}_i \gets \{x_i^j(0)\}_i - \{x_\ell^j(0)\} + \{x^j_\ell(T)\}$
                \Comment{Replace $x_\ell^j(0)$}
            \EndIf
        \EndFor
   \EndFor
   \State SourceSourceDANN($\{x_i^j\}_i$, $\{y_i^j\}_i$, $c_\sigma$, $r_\theta$, $d_\mu$) \Comment{Run DANN as usual.}
   \State Return trained model components
\end{algorithmic}
\end{algorithm}

\paragraph{Interpretation} In totality, this algorithm may be seen as employing a style of \textit{adversarial training} where, rather than generating examples to fool a task classifier -- e.g., the single-source DG approach of Volpi et al.~\citep{volpi2018generalizing} --, we instead generate examples to exploit the weaknesses of the feature extractor $r_\theta$ whose goal is to fool the domain discriminator. In this sense, the generated examples can be interpreted as \textit{cooperating} with the domain discriminator. Hence, we refer to the technique as DANN with Cooperative Examples, or DANNCE. For details on our implementation of DANNCE please see the pseudo-code in Algorithm Block~\ref{alg:example}. Additional details can also be found in Appendix~\ref{sec:appendix_b}.

%% file: 5_results.tex
In this section, we aim at addressing the primary point argued throughout this paper: the application of DANN to DG can benefit from (algorithmic) consideration of source diversity. To this end, our modus operandi is the comparison to recent state-of-the-art methods using a source-source DANN, or other domain alignment techniques, for domain generalization. See Appendix~\ref{sec:appendix_b} and code provided in supplement for all implementation details and additional experiments.  

\paragraph{Datasets and Hyper-Parameters}
We evaluate our method on two multi-source DG datasets. (1) PACS \citep{li2017deeper_PACS} contains 4  different styles of images (Photo, Art, Cartoon, and Sketch) with 7 common object categories. (2) Office-Home \citep{venkateswara2017deep} also contains 4 different styles of images (Art, Clipart, Product, and Real-W[orld]) with 65 common categories of daily objects. 
For both datasets, we follow standard experimental setups. We use 1 domain as target and the remaining 3 domains as sources. We report the average classification accuracy of the unseen target over 3 runs, using the model state at the last epoch to avoid peaking at the target. We select our hyper-parameters using leave-one-source-out CV \citep{balaji2018metareg}; this again avoids using the target in any way. Because some methods select parameters using a source train/val split, we use only the \textit{training} data of the standard splits for fairness. Other parameters of our setup, unrelated to our own method, are selected based on the environment of Matsuura and Harada~\citep{matsuura2019domain} (\textbf{MMLD}) -- a SOTA source-source DANN technique. For full details, see Appendix~\ref{sec:appendix_b}.

\paragraph{Our Models}
For the feature extractor $r_\theta$ we use AlexNet \citep{krizhevsky2012imagenet} for PACS and ResNet-18 \citep{he2016deep} for PACS and OfficeHome. Both are pretrained on ImageNet with the last fully-connected (FC) layer removed. For task classifier $c_\sigma$ and domain discriminator $d_\mu$ we use only FC layers. For \textbf{ERM} (often called \textit{Vanilla} or \textit{Deep All}) only $r_\theta$ and $c_\sigma$ are used and the model is trained on a mixture of all sources; this is a traditional DG baseline. For \textbf{DANN}, we add the domain discriminator $d_\mu$ and additionally update $r_\theta$ with $\mathcal{L}_{SD}$ (see Eq.~\eqref{eqn:sdann_obj}). Because we ultimately compare against DANN as a baseline, we must ensure our implementation is state-of-the-art. Therefore, we generally follow the implementation described by Matsuura and Harada~\citep{matsuura2019domain}, adding a commonly used Entropy Loss \citep{bengio1992optimization, shu2018dirt} and phasing-in the impact of $\mathcal{L}_{SD}$ on $r_\theta$ by setting $\lambda=2/(1+\exp(-\kappa\cdot p))-1$ in Eq.~\eqref{eqn:sdann_obj} with $p=\textrm{epoch}/\textrm{max\_epoch}$ and $\kappa = 10$. 

For our proposed method, \textbf{DANNCE}, we use the same baseline DANN, but update 50\% of the images (i.e., $\beta=0.5$) to \textit{cooperate} with the domain discriminator following Eq.~\eqref{eqn:update_rule_second_pass}. The number of update steps per image is 5 (i.e., $T=5$).

\begin{table}[t!]
\caption{\label{table:pacs_comparison}PACS and OfficeHome Results in Accuracy (\%). \textbf{avg}: Average of the target domain accuracies. \textbf{gain}: \textbf{avg} gain over the respective ERM (if reported).}
\resizebox{\columnwidth}{!}{%
\centering 
\footnotesize
\setlength{\tabcolsep}{4pt}
\renewcommand{\arraystretch}{1.10}
\begin{tabular}{lcccc|cc}
\specialrule{.1em}{.1em}{0.1em} 
\textbf{PACS} & \textbf{art} & \textbf{cartoon} & \textbf{sketch} & \textbf{photo} & \textbf{avg} & \textbf{gain} \\ \hline
\multicolumn{7}{c}{\textbf{AlexNet}}                                                                                                   \\ \hline
ERM           & 69.1         & 70.2             & 61.8            & 88.9           & 72.3         & -             \\
MMLD          & 68.5         & 72.2             & 66.3            & 89.3           & 74.1         & 1.8           \\
MMLD-2          & 67.0         & 70.6             & 67.8            & 89.4           & 73.7         & 1.4           \\
MMLD-3     & 69.3         & 72.8             & 66.4            & 89.0           & 74.4         & 2.1           \\ \hline
ERM           & 64.9         & 70.2             & 61.4            & 90.0           & 71.6         & -             \\
G2DM          & 66.6         & 73.4             & 66.2            & 88.1           & 73.6         & 2.0           \\ \hline
ERM           & 67.6         & 70.2             & 60.3            & 89.3           & 71.9         & -             \\
DANN          & 71.2         & 72.1             & 66.3            & 88.3           & 74.5         & 2.6           \\
DANNCE        & 70.9         & 72.0             & 67.9            & 89.6           & 75.1         & 3.2           \\
\hline\multicolumn{7}{c}{\textbf{ResNet-18}}                                                                                                    \\ \hline
ERM           & 77.0         & 75.9             & 69.2            & \multicolumn{1}{c|}{96.0}           & 79.5         & -             \\
MMD-AAE       & 75.2         & 72.7             & 64.2            & \multicolumn{1}{c|}{96.0}           & 77.0         & -2.5          \\
CrossGRAD     & 79.8         & 76.8             & 70.2            & \multicolumn{1}{c|}{96.0}           & 80.7         & 1.2           \\
DDAIG         & 84.2         & 78.1             & 74.7            & \multicolumn{1}{c|}{95.3}           & 83.1         & 3.6           \\ \hline
ERM           & 78.3         & 75.0             & 65.2            & \multicolumn{1}{c|}{96.2}           & 78.7         & -             \\
MMLD-2          & 81.3         & 77.2             & 72.3            & \multicolumn{1}{c|}{96.1}           & 81.8         & 3.1           \\
MMLD-3          & 79.6         & 76.8             & 71.2            & \multicolumn{1}{c|}{95.9}           & 80.9         & 3.1           \\ \hline
G2DM          & 77.8         & 75.5             & 77.6            & \multicolumn{1}{c|}{93.8}           & 81.2         & -             \\ \hline
ERM           & 78.1         & 75.6             & 66.0            & \multicolumn{1}{c|}{95.4}           & 78.7         & -             \\
DANN          & 80.2         & 77.6             & 70.0            & \multicolumn{1}{c|}{95.4}           & 80.8         & 2.0           \\
DANNCE        & 82.1         & 78.2             & 71.9            & \multicolumn{1}{c|}{94.7}           & 81.7         & 3.0    \\
\specialrule{.1em}{.1em}{0.1em}      
\textbf{OfficeHome} & \textbf{art} & \textbf{clipart} & \textbf{product} & \textbf{real-w} & \textbf{avg} & \textbf{gain} \\ \hline 
ERM                  & 58.9         & 49.4             & 74.3             & 76.2                & 64.7         & -             \\
MMD-AAE              & 56.5         & 47.3             & 72.1             & 74.8                & 62.7         & -2.0          \\
CrossGRAD            & 58.4         & 49.4             & 73.9             & 75.8                & 64.4         & -0.4          \\
DDAIG                & 59.2         & 52.3             & 74.6             & 76.0                & 65.5         & 0.8           \\ \hline
ERM                  & 60.0         & 49.0             & 75.4             & 76.8                & 65.3         & -             \\
DANN                 & 61.6             & 48.9                 & 75.8                 &  76.2                   &  65.6            & 0.3              \\
DANNCE               & 61.6             &  50.2                &  75.6                &  75.9                   & 65.8             & 0.5              \\ 
\specialrule{.1em}{.1em}{0.1em} 
\end{tabular}
}
\end{table}

\paragraph{Experimental Baselines} As mentioned, we focus on comparison to other methods proposing domain alignment for DG. Albuquerque et al.~\citep{albuquerque2019adversarial} (\textbf{G2DM}) and Li et al.~\citep{li2018domain_mmd} (\textbf{MMD-AAE}) propose variants of DANN,\footnote{MMD-AAE is based on the maximum-mean discrepancy \citep{gretton2012kernel} rather than $\mathcal{H}$-divergence.} and in particular, align domains by making updates to the feature extractor. As noted, Matsuura and Harada~\citep{matsuura2019domain} (\textbf{MMLD}) propose the DANN setup most similar to our baseline DANN. For MMLD, Matsuura and Harada~\citep{matsuura2019domain} additionally propose a source domain \textit{mixing} algorithm -- we denote this by MMLD-$K$ with $K$ the number of domains after re-clustering. Shankar et al.~\citep{shankar2018generalizing} (\textbf{CrossGRAD}) and Zhou et al.~\citep{zhou2020deep} (\textbf{DDAIG}), contrary to our work, generate examples which \textit{maximize} the domain loss. Because, they \textit{do not} update the feature extractor with the domain loss $\mathcal{L}_{SD}$ as we do, this may actually be viewed as domain-alignment by data generation (see Liu et al.~\citep{liu2019transferable} who first propose this technique). For MMD-AAE and CrossGRAD, we use results reported by Zhou et al.~\citep{zhou2020deep} because the original methods do not test on our datasets.

\paragraph{Analysis of Performance} Generally, in DG, the comparison of performance is subjective across different experimental setups -- a problem highlighted by a recent commentary on the experimental rigor of DG setups \citep{gulrajani2020search}. As such, we include reported results from other experimental setups, predominantly, to show our DANN implementation is a competitive baseline. This much is visible in Table~\ref{table:pacs_comparison}. For 2 out of 3 setups, our DANN \textit{alone} has higher overall accuracy than any other method.

Our focus, then, is the validation of our main argument using our strong DANN baseline. In this context, shown in Table~\ref{table:pacs_comparison}, ablation of DANNCE reveals substantial improvement upon the traditional source-source DANN in all PACS setups and marginal improvement in the OfficeHome setup. This is somewhat intuitive as OfficeHome has a staggering 65 categories to classify compared to 7 in PACS. Ultimately, the performance gains demonstrated by addition of DANNCE agrees with our main argument: increasing diversity when aligning domains can have practical benefits in DG.

\begin{figure}[t!]
    \centering
    \includegraphics[trim=13 13 10 0,clip,width=.93\columnwidth]{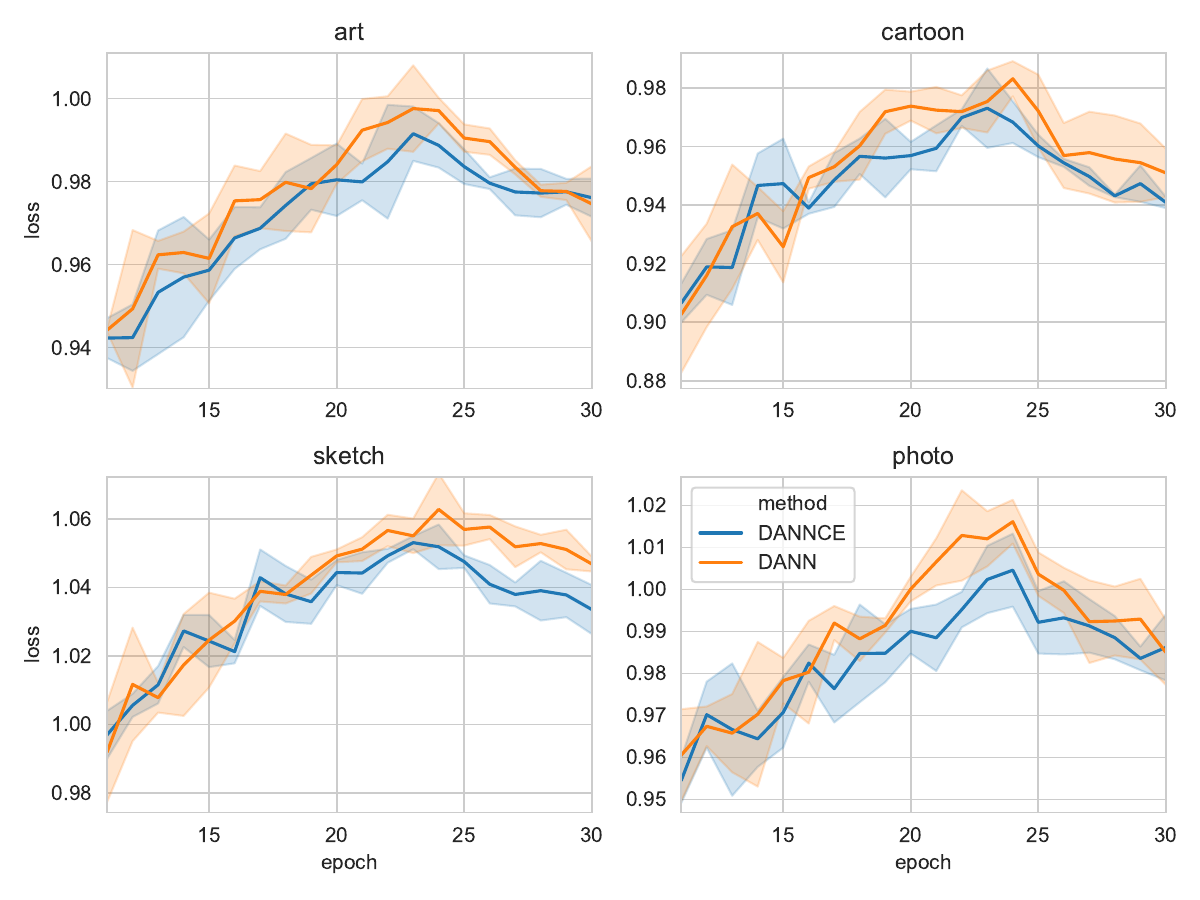}
    \caption{ Domain Discriminator Loss of DANN and DANNCE on PACS. For each target, we show the loss of its corresponding sources  during training.}
    \label{fig:domain_loss_alex}
\end{figure}

\begin{figure}[t!]
    \centering
    \includegraphics[trim=13 13 10 10,clip,width=.93\columnwidth]{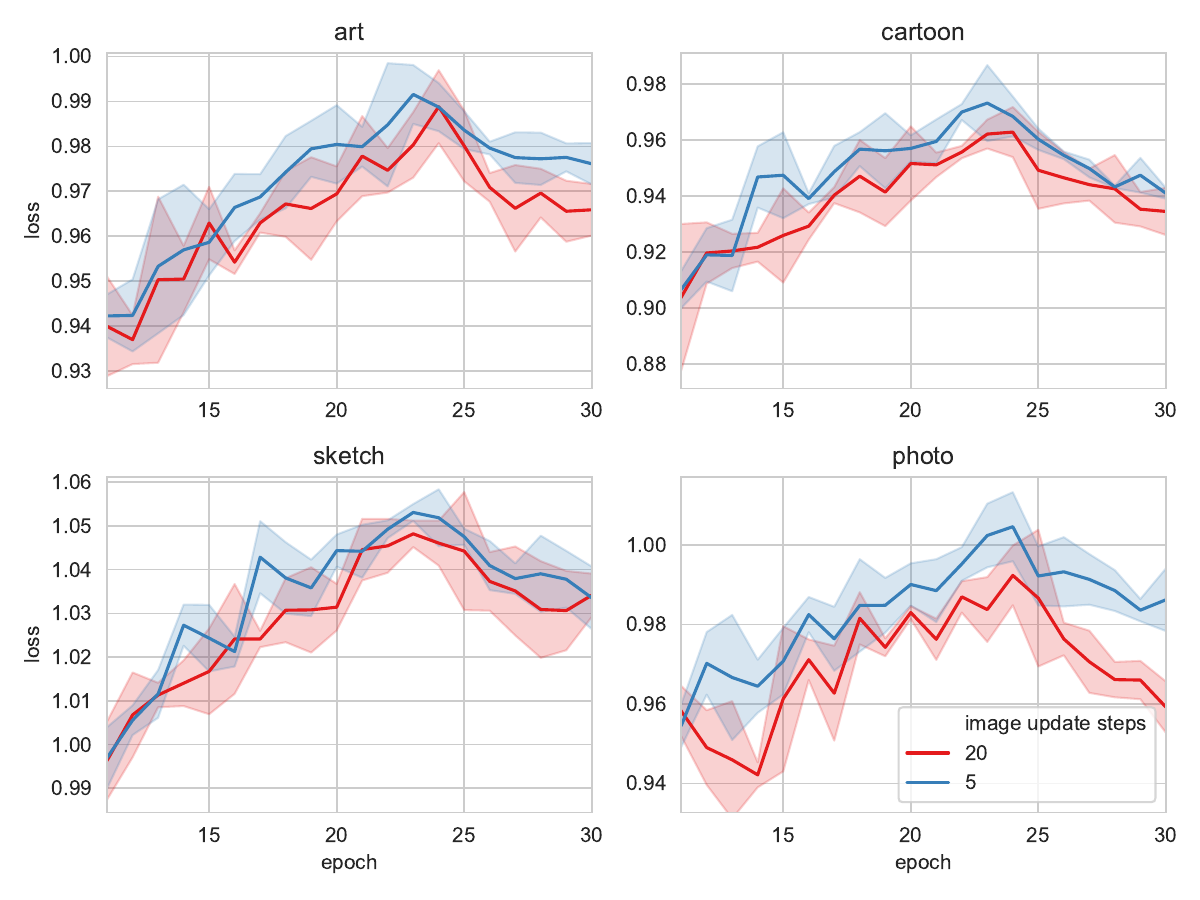}
    \caption{Domain Discriminator Loss of DANNCE with 5 and 20 Steps of Image Updates.}
    \label{fig:domain_loss_steps}
\end{figure}

\paragraph{Analysis of Loss Curves}
To measure domain diversity, we use the loss of the domain discriminator (averaged per epoch). This loss is used to proxy the $\mathcal{H}$-divergence (an inverse relationship). A lower loss should then indicate more domain diversity, and, has the benefit of \textit{dynamically} measuring diversity during training. Figure~\ref{fig:domain_loss_alex} shows the domain discriminator loss across epochs for our implementations of DANN and DANNCE using AlexNet on PACS. We generally see after epoch 15, the loss for DANNCE is lowest. Figure~\ref{fig:domain_loss_steps} further shows the effect of increasing the number of steps per image update. This suggests that increasing the number of updates has some control over the source domain diversity as intended. Finally, in both Figures, epochs 10 to 24 show the (inverted) smooth proxy for the domain divergence is increasing. This agrees with the formal claim made in Prop.~\ref{prop:contractions}. Although the trend changes after epoch 24, this is likely due to a decrease in $\gamma$ at this epoch, and thus, does not necessarily disagree with our formal claim. 


%% file: 6_related.tex
\paragraph{Domain Adaptation Theory}
Many works extend the theoretical framework of Ben-David et al. \cite{ben2010theory} to motivate new variants of DANN. Zhao et al.~\cite{zhao2018adversarial} consider the multi-source setting, Schoenauer-Sebag et al.~\cite{schoenauer2019multi} consider a multi-domain setting in which all domains have labels but large variability across domains must be handled, and Zhang et al.~\cite{zhang2019bridging} consider theoretical extensions to the multi-class setting using a margin-loss. Besides the theoretical perspective of Ben-David et al. in DA, there are many other works to consider. Mansour et al.~\cite{mansour2009domain} consider the case of general loss functions rather than the 01-error. Kuroki et al.~\cite{kuroki2019unsupervised} consider a domain-divergence which depends on the source-domain and, through this dependence, always produces a tighter bound. Many works also consider intergral probability metrics including: Redko et al.~\cite{redko2017theoretical}, Shen et al.~\cite{shen2018wasserstein}, and Johansson et al.~\cite{johansson2019support}. As has been discussed in this paper, the assumptions of various domain adaptation theories are of particular importance. Consequently, these assumptions are also important for DG. We discuss some assumptions in more detail in the next part. 
\paragraph{Assumptions in DA}
Ben-David et al.~\cite{david10a} show control of their divergence term as well as the ideal joint error $\lambda$ (so that both are small) give necessary and sufficient conditions for a large class of domain adaptation learners. These are the conditions which we control (in case of the divergence term) and assume (in case of the ideal joint error). Other assumptions for DA include the co-variate shift assumption in which the marginal feature distributions are assumed to change but the feature conditional label distributions across domains remain constant. As we have discussed, Zhao et al.~\cite{zhao2019learning} show that this assumption is not always enough in the context of DANN and  Johansson et al.~\cite{johansson2019support} provide similar conceptualizations. Still, this assumption can be useful in the context of model selection \cite{sugiyama2007covariate, you2019towards}. Another common assumption is label shift: the marginal label distributions disagree, but the label conditional feature distributions are the same. Again, this is related to the concern of Zhao et al.~\cite{zhao2019learning} since significant disagreement in the label distributions can cause DANN to fail miserably. Lipton et al.~\cite{lipton2018detecting} provide adaptation algorithms for this particular situation. Another assumption one can make for the benefit of algorithm design is the notion of \textit{generalized} label shift in which the label distributions may disagree and the feature conditional label distributions agree in an \textit{intermediate} feature space. As we have noted, Tachet des Combes, Zhao, et al.~\cite{combes_and_zhao} propose this assumption, devise new theoretical arguments under this assumptions, and suggest a number of algorithms based on their proposal.  
\paragraph{Domain Generalization Theory}
For DG, there is decidedly less theoretical work, but throughout our text, we have attempted to compare to the most relevant (and recent) -- a bound proposed by Albuquerque et al.~\cite{albuquerque2019adversarial}. Albeit, some different theoretical perspectives on DG do exist. In particular, Blanchard et al.~\cite{blanchard2011generalizing, blanchard2021domain} and Deng et al.~\cite{deng2020representation} consider domain generalization from the perspective of a meta-distribution which governs our observation of domains. Asymptotically, as we observe more domains, we can be more confident on the success of our algorithm. While this approach is interesting, our paper instead focuses on the case where we only have a relatively small number of domains from which to learn.
\paragraph{Algorithms in DG}
Besides DANN and other domain-aligning algorithms mentioned in this text, there are of course additional algorithmic perspectives on DG too. An early work in DG by Munadet et al.~\cite{muandet2013domain} proposes a kernel-based algorithm aimed at producing domain-invariant features with a strong theoretical justification. More recently, a common thread is the use of meta-learning (e.g. to simulate domain-holdout) seen in Li et al.~\cite{li2018learning_mldg}, Balaji et al.~\cite{balaji2018metareg}, and Dou et al.~\cite{dou2019domain}. Some authors, such as Wang et al.~\cite{wang2018learning} and Carlucci et al.~\cite{carlucci2019domain}, make additional assumptions on the domains to be seen and use this in algorithm design. Lastly, some works focus on the neural network components themselves, e.g., Li et al.~\cite{li2017deeper_PACS}. These architecture changes can be very effective (see Seo et al.~\cite{seo2019learning} for impressive results when modifying batch-normalization). Related to our paper's main point, we primarily focus on comparison to other methods proposing domain alignment for DG, especially those which are, in some sense, model agnostic. These additional references are discussed in \textit{\textbf{Experimental Baselines}}.

%% file: 7_conclusion.tex
In this work, we investigate the applicability of source-source DANN for domain generalization. Our theoretical results and interpretation suggest a complex relationship between the heterogeneity of the source domains and the usual process of domain alignment. Motivated by this, we construct an algorithmic extension for DANN which diversifies the sources via gradient-based image updates. Our empirical results and analyses support our findings.

One of the motivations of our algorithm is also one of the predominant limitations of our study. In particular, the behavior of DANN as a dynamic process is not well understood. Studying it as such can reveal to us new information. For example, in the proof of Proposition~\ref{prop:contractions}, we saw the importance of annealing the learning rate for DANN. We also use Proposition~\ref{prop:contractions} to motivate our algorithm design, but there are certainly open questions on the dynamic behavior of DANN and DANNCE. For example, it would be interesting to consider the competing objectives we have discussed in a more analytically tractable environment. Even for simple distributions, it is an open question as to how the hyper-parameters of DANNCE -- which intuitively balance the competing objectives -- may be optimally selected. On a related note, although we have assumed the ideal joint error is generally small, we have also pointed out that this is not always the case \cite{zhao2019learning}. While our promising results indicate this may not be an issue in practice, it is still interesting to consider this from a more theoretical perspective as well.
Finally, it is important to point out that our empirical investigation was limited to images. It is interesting to consider how our technique might extend to natural language or other areas where gradient-based algorithms are used for learning.

%% file: 8_appendix_a.tex
\subsection{On Theorem~\ref{thm:bendavid} in the Main Text}
This section covers much of the theoretical background our work relies on in detail. Statements by Ben-David et al.~\cite{ben2007analysis, ben2010theory} used to motivate the DANN algorithm \citep{ganin2015unsupervised} as well as statements on sample complexity \citep{kifer2004detecting} are included.
\subsubsection{Setup} We begin with a more detailed exposition of the setup assumed. We assume a space $\mathcal{X}$ and a class of deterministic hypotheses $\mathcal{H} \subseteq \{h \mid \mathcal{X} \to \{0,1\}\}$. In accordance with Ben-David et al.~\citep{ben2010theory}, for two functions $h$ and $f$ mapping from a space $\mathcal{X}$ into the set $\{0,1\}$, we define a disagreement measure with respect to a distribution $\mathbb{P}$ over $\mathcal{X}$ as below
\begin{equation}
\label{eqn:disagreement}
    \mathcal{E}_\mathbb{P}(h, f) = \mathbf{E}_{x \sim \mathbb{P}} \left\lvert h(x) - f(x) \right\rvert \\ 
    = \mathbf{E}_{x \sim \mathbb{P}} \left [\mathbb{I}[h(x) \neq f(x)]\right]
\end{equation}
where $\mathbb{I}$ is the indicator function; i.e., $\mathcal{E}_\mathbb{P}(h, f)$ is the probability that $h$ disagrees with $f$. If $h$ is a hypothesis and $f$ is a labeling function for $\mathbb{P}$ which we would like to approximate by $h$, we call this term the error of $h$ and write $\mathcal{E}_\mathbb{P}(h)$. We remark that the labeling function for $\mathbb{P}$ need not be in $\mathcal{H}$. Further, we sometimes permit the labeling function $f$ to have the continuous image $[0,1]$ to capture the possibility of a non-deterministic label. Lastly, for each distribution $\mathbb{P}$, we write an empirical sample as $\widehat{\mathbb{\mathbb{P}}}$, and generally, specify its size.

We also recall from the main text, the measure considered in this paper is based on the $\mathcal{H}$-divergence which itself is an adaptation of the $\mathcal{A}$-distance \cite{kifer2004detecting}. In particular, given two distributions $\mathbb{P}$, $\mathbb{Q}$ over a space $\mathcal{X}$ and a corresponding hypothesis class $\mathcal{H}$, the $\mathcal{H}$-divergence \citep{ben2010theory} is defined  
\begin{equation}
    d_\mathcal{H}(\mathbb{P}, \mathbb{Q}) = 2 \sup\nolimits_{h \in \mathcal{H}} \left \lvert \mathrm{Pr}_\mathbb{P}(I_h) - \mathrm{Pr}_\mathbb{Q}(I_h)\right \rvert
\end{equation}
where $I_h = \{x \in \mathcal{X} \mid h(x) = 1\}$.
To arrive at the $\mathcal{H}\Delta\mathcal{H}$-divergence,  Ben-David et al.~\citep{ben2010theory} define the symmetric difference hypothesis class $\mathcal{H}\Delta\mathcal{H}$. In particular, given a hypothesis class $\mathcal{H}$, the class $\mathcal{H} \Delta \mathcal{H}$ is the set of functions which are characteristic to disagreements between hypotheses. In details, we have 
\begin{equation}
\begin{split}
    g \in \mathcal{H} \Delta \mathcal{H} \quad \Leftrightarrow
    \quad g(x) = h_1(x) \oplus h_2(x)= \lvert h_1(x) - h_2(x) \rvert 
    \quad h_1, h_2 \in \mathcal{H}.
\end{split}
\end{equation}
Therefore, the $\mathcal{H}\Delta\mathcal{H}$-divergence is just a special case of the $\mathcal{H}$-divergence. As mentioned, this will be the measurement of divergence used in all considered bounds.

\subsubsection{Computing the $\mathcal{H}$-Divergence Empirically}
Here, we present Proposition \ref{prop:hdivemp}. This result is an important consideration for the design of the DANN algorithm.  In particular, both \cite{ben2007analysis, ben2010theory} and \cite{ganin2015unsupervised} suggest approximating the empirical $\mathcal{H}\Delta\mathcal{H}$-divergence by training a classifier to distinguish between the source and target distributions. To minimize the empirical $\mathcal{H}\Delta\mathcal{H}$-divergence, we should maximize this classifiers errors. Thus, this proposition can be viewed as motivation for our -- and many other authors' -- choice to approximate minimization of the divergence by maximization of a domain classifier's errors. 

\begin{proposition}[Ben-David et al.~\cite{ben2010theory} Lemma 2] 
\label{prop:hdivemp}
Provided a symmetric hypothesis class\footnote{A hypothesis class is symmetric if and only if for every $h \in \mathcal{H}$, we also have $1-h \in \mathcal{H}$.} and samples $\widehat{\mathbb{P}}$,
$\widehat{\mathbb{Q}}$ of size $n$

\begin{equation}
\begin{split}
\hat{d}_\mathcal{H}(\widehat{\mathbb{P}}, \widehat{\mathbb{Q}}) = 
2\left(1 - \min_{h \in \mathcal{H}}\left[ \frac{1}{n} \sum_{x \mid h(x) = 0} \mathbb{I}\left[x \in \widehat{\mathbb{P}} \right] 
+ \frac{1}{n} \sum_{x \mid h(x) = 1} \mathbb{I}\left[x \in \widehat{\mathbb{Q}} \right] \vphantom{\sum_{x \mid h(x) = 0}} \right] \vphantom{\sum_{x \mid h(x) = 0}} \right)
\end{split}
\end{equation}
\end{proposition}
\begin{proof}
We proceed in a similar fashion to \cite{ben2010theory}.  Let $h \in \mathcal{H}$ and consider the quantity
\begin{equation}
    1 - \left [ \frac{1}{n} \sum_{x \mid h(x) = 0} \mathbb{I}\left[x \in \widehat{\mathbb{P}} \right] + \frac{1}{n} \sum_{x \mid h(x) = 1} \mathbb{I}\left[x \in \widehat{\mathbb{Q}} \right] \right]. 
\end{equation}
We note two obvious facts. Every $x$ must belong to the sample $\widehat{\mathbb{Q}}$ or $\widehat{\mathbb{P}}$ and every $x$ must have $h(x) \in \{0,1\}$. Therefore, we can rewrite $1 = \frac{2n}{2n}$ and we have
\begin{equation}
\begin{split}
1 = 
& \frac{1}{2n}  \sum_{x \mid h(x) = 0} \left (\mathbb{I}\left[x \in \widehat{\mathbb{P}}\right] + \mathbb{I}\left[x \in \widehat{\mathbb{Q}} \right] \right ) 
+ \frac{1}{2n} \sum_{x \mid h(x) = 1} \left ( \mathbb{I}\left[x \in \widehat{\mathbb{P}}\right] + \mathbb{I}\left[x \in \widehat{\mathbb{Q}} \right]\right) 
\end{split}
\end{equation}
By taking the common denominator $2n$, we may then write
\begin{equation}
    \begin{split}
        & 1 - \left [ \frac{1}{n} \sum_{x \mid h(x) = 0} \mathbb{I}\left[x \in \widehat{\mathbb{P}} \right] + \frac{1}{n} \sum_{x \mid h(x) = 1} \mathbb{I}\left[x \in \widehat{\mathbb{Q}} \right] \right]\\
        & = \frac{1}{2n}  \sum_{x \mid h(x) = 0} \left (\mathbb{I}\left[x \in \widehat{\mathbb{Q}}\right] - \mathbb{I}\left[x \in \widehat{\mathbb{P}} \right] \right )
        +  \frac{1}{2n} \sum_{x \mid h(x) = 1} \left ( \mathbb{I}\left[x \in \widehat{\mathbb{P}}\right] - \mathbb{I}\left[x \in \widehat{\mathbb{Q}} \right]\right).
    \end{split}
\end{equation}
Now, for any sample $\widehat{\mathbb{P}}$ of size $n$, we have 
\begin{equation}
\mathrm{Pr}_{\widehat{\mathbb{P}}}(I_h) = \frac{1}{n} \sum_{x \mid h(x) = 1} \mathbb{I}\left[x \in \widehat{\mathbb{P}} \right] 
\end{equation}
and
\begin{equation}
    1 - \mathrm{Pr}_{\widehat{\mathbb{P}}}(I_h) = \frac{1}{n} \sum_{x \mid h(x) = 0} \mathbb{I}\left[x \in \widehat{\mathbb{P}} \right].
\end{equation}
Therefore, 
\begin{equation}
\begin{split}
& \frac{1}{2n}  \sum_{x \mid h(x) = 0} \left (\mathbb{I}\left[x \in \widehat{\mathbb{Q}}\right] - \mathbb{I}\left[x \in \widehat{\mathbb{P}} \right] \right )
+ \frac{1}{2n} \sum_{x \mid h(x) = 1} \left ( \mathbb{I}\left[x \in \widehat{\mathbb{P}}\right] - \mathbb{I}\left[x \in \widehat{\mathbb{Q}} \right]\right) \\
& = \frac{1}{2} \left ( 1 - \mathrm{Pr}_{\widehat{\mathbb{Q}}}(I_h) - (1 - \mathrm{Pr}_{\widehat{\mathbb{P}}}(I_h))\right) 
+ \frac{1}{2} \left ( \mathrm{Pr}_{\widehat{\mathbb{P}}}(I_h) - \mathrm{Pr}_{\widehat{\mathbb{Q}}}(I_h)\right) \\ 
& = \mathrm{Pr}_{\widehat{\mathbb{P}}}(I_h) - \mathrm{Pr}_{\widehat{\mathbb{Q}}}(I_h).
\end{split}
\end{equation}
Following the chain of inequalities and taking a maximum on both sides, we therefore have 
\begin{equation}
\begin{split}
    \max_{h \in \mathcal{H}} \mathrm{Pr}_{\widehat{\mathbb{P}}}(I_h) - \mathrm{Pr}_{\widehat{\mathbb{Q}}}(I_h)
    = 1 - \min_{h \in \mathcal{H}}\left [ \frac{1}{n} \sum_{x \mid h(x) = 0} \mathbb{I}\left[x \in \widehat{\mathbb{P}} \right]
    + \frac{1}{n} \sum_{x \mid h(x) = 1} \mathbb{I}\left[x \in \widehat{\mathbb{Q}} \right] \right].
\end{split}
\end{equation}
Finally, to complete the proof, we note that
\begin{equation}
    \mathrm{Pr}_{\widehat{\mathbb{P}}}(I_h) - \mathrm{Pr}_{\widehat{\mathbb{Q}}}(I_h) = \mathrm{Pr}_{\widehat{\mathbb{Q}}}(I_{1-h}) - \mathrm{Pr}_{\widehat{\mathbb{P}}}(I_{1-h}).
\end{equation}
Since $\mathcal{H}$ is assumed symmetric, we therefore have 
\begin{equation}
\begin{split}
    \max_{h \in \mathcal{H}} \mathrm{Pr}_{\widehat{\mathbb{P}}}(I_h) - \mathrm{Pr}_{\widehat{\mathbb{Q}}}(I_h)
    = \max_{h \in \mathcal{H}} \left \lvert \mathrm{Pr}_{\widehat{\mathbb{P}}}(I_h) - \mathrm{Pr}_{\widehat{\mathbb{Q}}}(I_h) \right \rvert
\end{split}
\end{equation}
and we are done. 
\end{proof}

\subsubsection{Proof of Theorem~\ref{thm:bendavid}}
\label{sec:bendavid}
Here, we present Theorem 2.1 of the main text (referenced in the Appendix as Theorem \ref{thm:bendavid}). We begin with a required Lemma for the final proof.
\begin{lemma}[\cite{ben2010theory} Lemma 3] 
\label{lem:bendavid}
Let $\mathcal{X}$ be a space and $\mathcal{H}$ a class of hypotheses corresponding to this space. Let $\mathbb{P}$ and $\mathbb{Q}$ be distributions over $\mathcal{X}$. Then for any hypotheses $h_1, h_2 \in \mathcal{H}$, we have 
\begin{equation}
       \left \lvert \mathcal{E}_\mathbb{P}(h_1, h_2) - \mathcal{E}_\mathbb{Q}(h_1, h_2)\right \rvert \leq \frac{1}{2} d_{\mathcal{H} \Delta \mathcal{H}}(\mathbb{P}, \mathbb{Q}) 
\end{equation}
\end{lemma}
\begin{proof}
We proceed in a similar fashion to \cite{ben2010theory}. By definition of the $\mathcal{H}$-divergence we have
\begin{equation}
\begin{split}
    d_{\mathcal{H} \Delta \mathcal{H}}(\mathbb{P}, \mathbb{Q})  
    & = 2 \sup_{g \in \mathcal{H} \Delta \mathcal{H}} \left \lvert \mathrm{Pr}_\mathbb{P}(I_g) - \mathrm{Pr}_\mathbb{Q}(I_g)\right \rvert \\
    &  = 2 \sup_{h, h^\prime \in \mathcal{H}} \lvert \mathrm{Pr}_\mathbb{P}(\{ x \mid h(x) \neq h^\prime(x)\})
     - \mathrm{Pr}_\mathbb{Q}(\{ x \mid h(x) \neq h^\prime(x)\}) \rvert \\
    &  = 2 \sup_{h, h^\prime \in \mathcal{H}}  \lvert \mathbf{E}_{x \sim \mathbb{P}} \left [ \mathbb{I}[h(x) \neq h^\prime(x)]\right]
    - \mathbf{E}_{x \sim \mathbb{Q}} \left [ \mathbb{I}[h(x) \neq h^\prime(x)]\right] \rvert \\
    & = 2 \sup_{h, h^\prime \in \mathcal{H}} \left \lvert\mathcal{E}_\mathbb{P}(h, h^\prime) - \mathcal{E}_\mathbb{Q}(h, h^\prime)\right \rvert  \\
    & \geq 2 \left \lvert\mathcal{E}_\mathbb{P}(h_1, h_2) - \mathcal{E}_\mathbb{Q}(h_1, h_2)\right \rvert.
\end{split}
\end{equation}
Here, the second equality follows directly from the definition of $\mathcal{H} \Delta \mathcal{H}$, the fourth equality follows from Main Text Eq.~6, and the last inequality follows by properties of the supremum.
\end{proof}

Using Lemma \ref{lem:bendavid}, we may present the proof of Theorem \ref{thm:bendavid}. Our statement is modified, omitting empirical quantities. We invite the reader to view Theorem 2 of \cite{ben2010theory} for the full result.
\begin{proof}
We proceed in a similar fashion to \cite{ben2010theory}. First, note the triangle inequality of classification error  \citep{crammer2007learning, ben2007analysis} which states that given \textit{any} labeling functions $h_1, h_2, h_3$, we have
\begin{equation}
    \mathcal{E}_\mathbb{A}(h_1, h_2) \leq \mathcal{E}_\mathbb{A}(h_1, h_3) + \mathcal{E}_\mathbb{A}(h_2, h_3).
\end{equation}
Then, let $h \in \mathcal{H}$, let $\eta = \mathrm{arg}\min_{h\in\mathcal{H}} \mathcal{E}_\mathbb{Q}(h) + \mathcal{E}_\mathbb{P}(h)$, and let $f_\mathbb{P}, f_\mathbb{Q}$ be the true labeling functions of $\mathbb{P}, \mathbb{Q}$ on $\mathcal{X}$, respectively. Given this, we have
\begin{equation}
\begin{split}
    \mathcal{E}_\mathbb{Q}(h)
    & = \mathcal{E}_\mathbb{Q}(h, f_\mathbb{Q}) \\
    & \leq \mathcal{E}_\mathbb{Q}(\eta, f_\mathbb{Q}) + \mathcal{E}_\mathbb{Q}(\eta, h) \\
    & \leq \mathcal{E}_\mathbb{Q}(\eta, f_\mathbb{Q}) + \mathcal{E}_\mathbb{P}(\eta, h) + \left \lvert \mathcal{E}_\mathbb{P}(\eta, h) - \mathcal{E}_\mathbb{Q}(\eta, h)\right \rvert \\
    & \leq \mathcal{E}_\mathbb{Q}(\eta, f_\mathbb{Q}) + \mathcal{E}_\mathbb{P}(\eta, h) + \frac{1}{2} d_{\mathcal{H} \Delta \mathcal{H}}(\mathbb{Q}, \mathbb{P}) \\
    & \leq \mathcal{E}_\mathbb{Q}(\eta, f_\mathbb{Q}) + \mathcal{E}_\mathbb{P}(\eta, f_\mathbb{P}) + \mathcal{E}_\mathbb{P}(h, f_\mathbb{P}) 
    + \frac{1}{2} d_{\mathcal{H} \Delta \mathcal{H}}(\mathbb{Q}, \mathbb{P}) \\ 
    & = \mathcal{E}_\mathbb{Q}(\eta) + \mathcal{E}_\mathbb{P}(\eta) + 
    \mathcal{E}_\mathbb{P}(h) +
    \frac{1}{2} d_{\mathcal{H} \Delta \mathcal{H}}(\mathbb{Q}, \mathbb{P}).
\end{split}
\end{equation}
Here, the second inequality comes from considering both the cases $\mathcal{E}_\mathbb{B}(\eta, h) > \mathcal{E}_\mathbb{A}(\eta, h)$ and $\mathcal{E}_\mathbb{A}(\eta, h) > \mathcal{E}_\mathbb{B}(\eta, h)$, the third inequality comes from Lemma \ref{lem:bendavid}, and all other inequalities follow from the triangle inequality of classification error.
\end{proof}

\subsubsection{Sample Complexity of Theorem~\ref{thm:bendavid}}
\label{sec:sample_complexity}
Here, we present Proposition \ref{prop:generror}. This Proposition contributes the main result required to derive generalization bounds for Theorem~\ref{thm:bendavid}. Since Theorem~\ref{thm:bendavid} is modified from Theorem 2 of Ben-David et al.~\citep{ben2010theory}, we direct the reader to this proof for further details. 

\paragraph{Remark on Sample Complexity} In general, we choose to omit discussion of sample complexity from the main text. In the usual case, where $\mathcal{H}$ is a class of neural networks, the VC Dimension \citep{vapnik1999nature} is usually larger\footnote{For NNs, it is $O(WL \log W)$ where $W$/$L$ are the number of weights/layers \citep{bartlett2019nearly}.} than the number of samples. As can be seen in the statement of Proposition \ref{prop:generror}, this causes problems in the interpretation and assumptions of the generalization bound. Despite this fact, Ganin and Lempitsky~\citep{ganin2015unsupervised} have shown that (empirically) this is a non-issue for application of DANN. With this said, some readers may rightly desire tighter bounds on empirical quantities when dealing with neural networks. Recently, some works have shown success in deriving much tighter bounds on empirical quantities (like error) for stochastic neural networks using the PAC-Bayesian framework Dziugaite et al.~\citep{dziugaite2017computing}. Within the PAC-Bayesian framework, Germain et al.~\citep{germain2020pac} provide a distribution divergence psuedometric very similar to the $\mathcal{H}\Delta\mathcal{H}$-divergence. As mentioned in the main text, the important property we use in our results is the psuedometric property, so we expect our results to hold in this more modern formulation as well. 

In any case, we remark that generalization bounds can be derived for Theorem~\ref{thm:bendavid} and other results in this paper by application of the below statement. For a more detailed discussion, where empirical quantities are considered and generalization bounds are discussed in a variety of circumstances, we direct the reader to the original work of Ben-David et al.~\citep{ben2010theory}.

\begin{proposition}[Ben-David et al.~\citep{ben2010theory} Lemma 2; Kifer et al.~\citep{kifer2004detecting} Theorem 3.2]
\label{prop:generror}
Let $\mathcal{X}$ be a space and $\mathcal{H}$ be a class of hypotheses corresponding to this space with VC dimension $d$. Suppose $\mathbb{P}$ and $\mathbb{Q}$ are distributions over $\mathcal{X}$ with corresponding samples $\widehat{\mathbb{P}}$ and $\widehat{\mathbb{Q}}$ of size $n$. Suppose $\hat{d}_\mathcal{H}(\widehat{\mathbb{P}}, \widehat{\mathbb{Q}})$ is the empirical $\mathcal{H}$-divergence between samples. Then, for any $\delta \in (0,1)$ the following holds with probability at least $1-\delta$
\begin{equation}
\begin{split}
    d_\mathcal{H}(\mathbb{P}, \mathbb{Q})  \leq \hat{d}_\mathcal{H}(\widehat{\mathbb{P}}, \widehat{\mathbb{Q}}) + O\left(\sqrt{\tfrac{d \log (\frac{2n}{d}) + \log (\frac{4}{\delta})}{n}}\right)
\end{split}
\end{equation}
\end{proposition}

\newpage
\subsection{On the $\mathcal{H}\Delta\mathcal{H}$-divergence with Comparison to a PAC-Bayesian Distribution Divergence}
Here, we prove some useful facts about the $\mathcal{H}\Delta\mathcal{H}$-divergence. In fact, these are the essential properties used to prove our formal claims in the main text. Most of these are known and have been used by other authors, but we restate and prove them here for completeness. One important point of this discussion is to demonstrate the relation between a second distribution divergence proposed by Germain et al.~\citep{germain2020pac} within the PAC-Bayes framework. As we will show, this PAC-Bayesian divergence exhibits the same properties. The consequence is that much of our formal discussion holds for this more modern divergence as well.

\subsubsection{A Nice Property of Mixture Distributions}
Below we provide a nice property of mixture distributions when considering their divergence. We are aware of variants of this result which have been observed by both Zhao et al.~\citep{zhao2018adversarial} and Albuquerque et al.~\citep{albuquerque2019adversarial} in derivation of their own bounds. We use this result in most of our proofs involving mixtures.

\begin{lemma}
\label{lem:mixtures_are_bounded_above_by_sums}
Let $\mathcal{X}$ be a space and let $\mathcal{H}$ be a class of hypotheses corresponding to this space. Let the collection $\{\mathbb{P}_i\}_{i=1}^k$ be distributions over $\mathcal{X}$. Now, suppose also that $\mathbb{Q}$ is a mixture of the component distributions $\{\mathbb{P}_i\}_i$; that is, for any set $A$,  we have $\mathrm{Pr}_\mathbb{Q}(A) = \sum_i \alpha_i \mathrm{Pr}_{\mathbb{P}_i}(A)$ with $\sum_i \alpha_i = 1$ and $\alpha_i \geq 0, \forall i$. Then, for any distribution $\mathbb{P}$, the following holds
\begin{equation}
    d_{\mathcal{H}\Delta\mathcal{H}}(\mathbb{P}, \mathbb{Q}) \leq \sum\nolimits_i \alpha_i  d_{\mathcal{H}\Delta\mathcal{H}}(\mathbb{P}, \mathbb{P}_i).
\end{equation}
\end{lemma}
\begin{proof} The result follows from the chain below
\begin{equation}
\label{eqn:chain_for_lemma}
\begin{split}
& d_{\mathcal{H}\Delta\mathcal{H}}(\mathbb{P}, \mathbb{Q}) = 2 \sup_{h \in \mathcal{H}\Delta\mathcal{H}} \left \lvert\mathrm{Pr}_{\mathbb{P}}(I_h) - \mathrm{Pr}_{\mathbb{Q}}(I_h) \right \rvert \\
& = 2 \sup_{h \in \mathcal{H}\Delta\mathcal{H}} \left \lvert\mathrm{Pr}_{\mathbb{P}}(I_h) - \sum\nolimits_j \alpha_j\mathrm{Pr}_{\mathbb{P}_j}(I_h) \right \rvert \\
& = 2 \sup_{h \in \mathcal{H}\Delta\mathcal{H}} \left \lvert\sum\nolimits_j \alpha_j \mathrm{Pr}_{\mathbb{P}}(I_h) - \sum\nolimits_j \alpha_j\mathrm{Pr}_{\mathbb{P}_j}(I_h) \right \rvert \\
& = 2 \sup_{h \in \mathcal{H}\Delta\mathcal{H}} \left \lvert\sum\nolimits_j \alpha_j \left ( \mathrm{Pr}_{\mathbb{P}}(I_h) - \mathrm{Pr}_{\mathbb{P}_j}(I_h)\right) \right \rvert \\
& \leq 2\sum\nolimits_j \alpha_j  \sup_{h \in \mathcal{H}\Delta\mathcal{H}} \left \lvert\mathrm{Pr}_{\mathbb{P}}(I_h) - \mathrm{Pr}_{\mathbb{P}_j}(I_h) \right \rvert \\
& = \sum\nolimits_j \alpha_j  d_{\mathcal{H}\Delta\mathcal{H}}(\mathbb{P}, \mathbb{P}_j).
\end{split}
\end{equation}
Here, the results follow mostly by definition or arithmetic, but we highlight some exceptions. The third equality follows because the coefficients $\{\alpha_j\}_j$ sum to 1. The only inequality follows by application of the triangle inequality (for absolute values) and properties of the supremum. In particular, for any $h^* \in \mathcal{H}\Delta\mathcal{H}$, we have
\begin{equation}
\begin{split}
& \sum\nolimits_j \alpha_j  \sup_{h \in \mathcal{H}\Delta\mathcal{H}} \left \lvert\mathrm{Pr}_{\mathbb{P}}(I_h) - \mathrm{Pr}_{\mathbb{P}_j}(I_h) \right \rvert \geq \sum\nolimits_j \alpha_j \left \lvert\mathrm{Pr}_{\mathbb{P}}(I_{h^*}) - \mathrm{Pr}_{\mathbb{P}_j}(I_{h^*}) \right \rvert \\
& \geq \left \lvert\sum\nolimits_j \alpha_j \left ( \mathrm{Pr}_{\mathbb{P}}(I_{h^*}) - \mathrm{Pr}_{\mathbb{P}_j}(I_{h^*})\right) \right \rvert
\end{split}
\end{equation}
where the first inequality follows because the supremum is defined as an upper-bound and the second follows from the triangle inequality. But, the supremum is also specified as the \textit{least} upper bound, so with 
\begin{equation}
\sum\nolimits_j \alpha_j  \sup_{h \in \mathcal{H}\Delta\mathcal{H}} \left \lvert\mathrm{Pr}_{\mathbb{P}}(I_h) - \mathrm{Pr}_{\mathbb{P}_j}(I_h) \right \rvert \geq \left \lvert\sum\nolimits_j \alpha_j \left ( \mathrm{Pr}_{\mathbb{P}}(I_{h^*}) - \mathrm{Pr}_{\mathbb{P}_j}(I_{h^*})\right) \right \rvert
\end{equation}
for any $h^* \in \mathcal{H}\Delta\mathcal{H}$, we may use the \textit{least} upper bound property of $\sup_{h \in \mathcal{H}\Delta\mathcal{H}} \left \lvert\sum\nolimits_j \alpha_j \mathrm{Pr}_{\mathbb{P}_i}(I_h) - \sum\nolimits_j \alpha_j\mathrm{Pr}_{\mathbb{P}_j}(I_h) \right \rvert$ to observe that 
\begin{equation}
\begin{split}
  & \sum\nolimits_j \alpha_j  \sup_{h \in \mathcal{H}\Delta\mathcal{H}} \left \lvert\mathrm{Pr}_{\mathbb{P}}(I_h) - \mathrm{Pr}_{\mathbb{P}_j}(I_h) \right \rvert \\
  & \geq \sup_{h \in \mathcal{H}\Delta\mathcal{H}} \left \lvert\sum\nolimits_j \alpha_j \mathrm{Pr}_{\mathbb{P}}(I_h) - \sum\nolimits_j \alpha_j\mathrm{Pr}_{\mathbb{P}_j}(I_h) \right \rvert
\end{split}
\end{equation}
affirming the third inequality.
\end{proof}

\subsubsection{Triangle Inequality}
Below, we provide proof that the $\mathcal{H}\Delta\mathcal{H}$-divergence abides by the triangle inequality. This result is used by many authors, although we have not seen proof. As noted in the main text, this (along with some other easy to verify properties) make the $\mathcal{H}\Delta\mathcal{H}$-divergence a \textit{psuedometric}. We use the triangle inequality a great deal throughout our proofs.

\begin{proposition}
\label{prop:triangle_ineq}
Let $\mathcal{H}$ be and arbitrary class of hypotheses over $\mathcal{X}$. Then, the $\mathcal{H}$-divergence abides by the triangle-inequality.
\end{proposition}
\begin{proof}
Let $\mathbb{P}$, $\mathbb{Q}$, and $\mathbb{S}$ be distributions over $\mathcal{X}$. We observe the following chain of inequalities
\begin{equation}
\begin{split}
    & d_{\mathcal{H}}(\mathbb{P}, \mathbb{Q}) = 2 \sup_{h \in \mathcal{H}} \left \lvert\mathrm{Pr}_{\mathbb{P}}(I_h) - \mathrm{Pr}_{\mathbb{Q}}(I_h) \right \rvert \\
    & = 2 \sup_{h \in \mathcal{H}} \left \lvert\mathrm{Pr}_{\mathbb{P}}(I_h) - \mathrm{Pr}_{\mathbb{S}}(I_h) + \mathrm{Pr}_{\mathbb{S}}(I_h) - \mathrm{Pr}_{\mathbb{Q}}(I_h) \right \rvert \\
    & \leq 2 \sup_{h \in \mathcal{H}} \left \lvert\mathrm{Pr}_{\mathbb{P}}(I_h) - \mathrm{Pr}_{\mathbb{S}}(I_h) \right \rvert + 2 \sup_{h \in \mathcal{H}} \left \lvert\mathrm{Pr}_{\mathbb{S}}(I_h) - \mathrm{Pr}_{\mathbb{Q}}(I_h) \right \rvert \\ & 
    = d_{\mathcal{H}}(\mathbb{P}, \mathbb{S}) + d_{\mathcal{H}}(\mathbb{S}, \mathbb{Q}).
\end{split}
\end{equation}
The inequality follows from an argument similar to that used in defense of Lemma~\ref{lem:mixtures_are_bounded_above_by_sums}, Equation~\ref{eqn:chain_for_lemma}.
\end{proof}

\subsubsection{Comparison to the Domain Disagreement \cite{germain2020pac}}
The domain disagreement is another distribution divergence proposed by Germain et al.~\citep{germain2020pac}. As noted by the these authors, the divergence is in fact designed as the PAC-Bayesian analog of the $\mathcal{H} \Delta \mathcal{H}$-divergence. We define the \textit{domain disagreement} below for a distribution $\rho$ over $\mathcal{H}$
\begin{equation}
    \mathrm{dis}_\rho(\mathbb{P}, \mathbb{S}) = \left \lvert \mathbf{E}_{(h,h')\sim \rho^2}  \ \left [\mathcal{E}_\mathbb{P}(h, h') - \mathcal{E}_\mathbb{S}(h, h')\right ]\right \rvert.
\end{equation}

As it turns out, the domain disagreement abides by a triangle-inequality and further satisfies Lemma~\ref{lem:mixtures_are_bounded_above_by_sums}. The former is a simple consequence of the fact that the domain disagreement is also a \textit{pseudometric} \cite{germain2020pac}. The latter is not so trivial to see, but we provide a quick sketch of the required steps below. Assuming $\mathbb{S}$ is a mixture as in Lemma~\ref{lem:mixtures_are_bounded_above_by_sums} we have 
\begin{equation}
    \begin{split}
        \mathrm{dis}_\rho(\mathbb{P}, \mathbb{S})  & = \left \lvert \mathbf{E}_{(h,h')\sim \rho^2}  \ \left [\mathcal{E}_\mathbb{P}(h, h') - \mathcal{E}_\mathbb{S}(h, h')\right ]\right \rvert \\
        & = \left \lvert \mathbf{E}_{(h,h')\sim \rho^2}  \ \left [\mathcal{E}_\mathbb{P}(h, h') - \sum\nolimits_i \alpha_i \mathcal{E}_{\mathbb{P}_i}(h, h')\right ]\right \rvert \\
        & = \left \lvert \mathbf{E}_{(h,h')\sim \rho^2}  \ \left [\sum\nolimits_i \alpha_i \left (\mathcal{E}_\mathbb{P}(h, h') -  \mathcal{E}_{\mathbb{P}_i}(h, h') \right )\right ]\right \rvert \\
        & = \left \lvert \sum\nolimits_i \alpha_i \mathbf{E}_{(h,h')\sim \rho^2}  \ \left [\mathcal{E}_\mathbb{P}(h, h') - \mathcal{E}_{\mathbb{P}_i}(h, h')\right ]\right \rvert \\
        & \leq \sum\nolimits_i \alpha_i \left \lvert  \mathbf{E}_{(h,h')\sim \rho^2}  \ \left [\mathcal{E}_\mathbb{P}(h, h') - \mathcal{E}_{\mathbb{P}_i}(h, h')\right ]\right \rvert.
    \end{split}
\end{equation}
The steps above generally follow from arithmetic similar to Eq.~\eqref{eqn:chain_for_lemma} or by linearity of the expectation. The last inequality uses properties of the absolute value. 

Harking back to the main point, we remind the reader that Lemma~\ref{lem:mixtures_are_bounded_above_by_sums} and the triangle-inequality the primary tools needed for our results. As such, much of our formal discussion holds for this more modern divergence as well.

%% file: 9_appendix_b.tex
We provide full details of our experiments and implementation. In the supplementary material package, \textbf{we also provide the fully functioning implementation of our approach}. Scripts for notable experimental setups and associated dataset links are provided for reproducibility. These will be publicly available on github upon acceptance.

\subsection{Datasets}
\label{sec:appendix_datasets_full}
\paragraph{PACS} PACS is a relatively new domain generalization dataset based on different styles of images \cite{li2017deeper_PACS}. Across 4 domains (Photo, Art Painting, Cartoon, and Sketch), there are 7 common object categories: dog, elephant, giraffe, guitar, horse, house, and person. There are a total of 9991 images. We split each domain into 90\% for training set and 10\% for validation set. The detail of PACS splits can be found in our codebase ({\tt /paths/PACS/}).

\paragraph{OfficeHome} Office-Home \citep{venkateswara2017deep} also contains 4 different styles of images (Art, Clipart, Product, and Real-W[orld]) with 65 common categories of daily objects. We use the down-sampled and preprocessed dataset curated by Zhou et al.~\cite{zhou2020deep} who propose DDAIG (compared to in the main text).

\subsection{Training details}
\label{sec:appendix_training_full}
As mentioned in the main text, our setup closely follows \cite{matsuura2019domain}. We implement our model in PyTorch \citep{pytorch_NEURIPS2019_9015}. Our model was trained by using 30 epochs, batch size 128
, and SGD optimizer configured with momentum 0.9 and weight decay 5e-4. We augment images using the same strategy as Matsuura and Harada~\cite{matsuura2019domain}, employing color-jitter, horizontal flips, and cropping. For PACS (AlexNet and ResNet), the initial learning rate of the feature extractors are 1e-3. The classifiers (i.e., task classifier and domain discriminator) have 10 times the initial learning rates of the corresponding feature extractor's initial learning rate since they are not pre-trained. We decrease all learning rates by a factor of 10 after 24 epochs. All losses are weighted equally by default, but as mentioned, following Matsuura and Harada~\cite{matsuura2019domain}, we phase-in the impact of $\mathcal{L}_{SD}$ and the entropy loss on $r_\theta$ by using the weight $\lambda=2/(1+\exp(-\kappa\cdot p))-1$. For OfficeHome, experimental settings are almost identical. We deviate only slightly by lowering the learning rate of the domain-discriminator by a factor of 10 and lowering the magnitude of $\mathcal{L}_{SD}$ by a factor of 4 when updating the feature extractor.\footnote{Studying learning curves early-on during experimentation suggested that (1) the domain-discriminator was learning to quickly for the feature extractor and (2) the feature extractor was ignoring the classification loss and focusing on $\mathcal{L}_{SD}$ as a result. These changes aimed to accommodate both of these observations and generally improved the learning curves.}

All experiments were run on an NVIDIA GeForce RTX 2080 Ti GPU 11GB. We used the helpful Weights and Biases tool \citep{wandb} during experimentation for visualizing our model training and results.

\subsection{Network Architectures}\label{sec:appendix_archtiectures_full}
In this section, we provide details of the network architectures of our model components.

\paragraph{Feature Extractors} 
We use AlexNet \citep{krizhevsky2012imagenet} for PACS and ResNet-18 \citep{he2016deep} for PACS and OfficeHome. In both cases, we pretrain on ImageNet \citep{imagenet_cvpr09} with the last FC layer removed. We note that we used a Caffe version of AlexNet implemented in PyTorch to follow related recent works \citep{carlucci2019domain,matsuura2019domain} which showed consistently competitive Deep All baseline accuracies. The exact implementations can be found in our codebase ({\tt /src/models/caffenet/models.py} and {\tt /src/models/resnet.py}). The pretrained model for AlexNet is also included in our supplementary materials. For ResNet-18, it is loaded from \texttt{torchvision} in the code.

\paragraph{Classifiers}
The class classifier for is a single fully connected (FC) layer. The domain discriminator follows the design of Matsuura and Harada~\cite{matsuura2019domain} and is a simple stack of fully connected layers:
\begin{equation}\footnotesize
    \texttt{FC(1024)-ReLU-DropOut(0.5)-FC(1024)-ReLU-DropOut(0.5)-FC(num\_of\_class)}
\end{equation} Again following \cite{matsuura2019domain}, the class classifier has an xavier (glorot) uniform initialization \citep{glorot2010understanding} with gain set to $0.1$, while the domain classifier uses the PyTorch \citep{pytorch_NEURIPS2019_9015} default initialization (version 1.4). The exact implementation can be found in the code, specifically module {\tt /src/models}.

\subsection{Hyper-Parameters of DANNCE} While we generally try to follow Matsuura and Harada~\cite{matsuura2019domain} as closely as possible to ensure our baseline DANN is state-of-the-art, we cannot use existing hyper-parameter choices for our novel algorithm (i.e., DANNCE). To perform the image updates (Line~5, Algorithm~1) we use the Adam optimizer \citep{kingma2014adam}. Generally, we fix $\beta=0.5$ and $t=5$ in Algorithm~1. To maintain realistic image values, we clamp the pixel-values of the resulting images after each update based on the max and min pixel values of the PACS dataset. Yosinski et al.~\cite{yosinskiunderstanding} also use image-space gradient updates and further identify the addition of Gaussian blurring to be an important parameter for producing realistic images. Based on one of the optimal settings described by Yosinski et al.~\cite{yosinskiunderstanding}, we use Gaussian blurring once every 4 steps of our algorithm. We provide ablation of the effect of blurring in Table~\ref{table:pacs_ablation} which reveals that blurring may indeed be important for our method when applied to images, but \textit{importantly} also shows that our gain in performance does not \textit{only} come from blurring because blurring without our algorithm actually hurts performance of the baseline DANN. 

We select the learning-rate and weight-decay of our Adam optimizer from the set $\{$1e-2, 1e-2, 1e-3$\}$ based on a leave-one-source-out cross validation used by Balaji et al.~\cite{balaji2018metareg}. For example, when Art is the unseen target, we run DANNCE with every parameter setting holding out each of Cartoon, Photo, and Sketch as a \textit{pseudo} target domain -- the pair of parameters performing best on the simulated holdout task (averaged over the three \textit{psuedo} targets) is used for a final training phase including all three sources and evaluated on Art. Note that this does not use the unseen target (e.g., Art) at all in the hyper-parameter selection, mimicking a real-life DG scenario. For PACS (AlexNet and ResNet-18), we run this leave-one-source-out cross validation. In the interest of time, for OfficeHome, we use the best parameter setting of PACS AlexNet (averaged over \textit{all} pseudo targets) instead of performing a leave-one-source-out cross validation. Clearly, this still does not use any information from the unseen target. The exact parameter settings in every case may be found in the code within each experiment's bash script (see directory {\tt final\_scripts}).

\begin{table}[t!]
\caption{\label{table:pacs_ablation}Ablation of Gaussian Blurring for DANNCE on PACS AlexNet. \textbf{gain} is computed with respect to the baseline DANN. For \textit{DANN with BLUR}, we use the same blurring schedule as used in our DANNCE implementation, but make no image updates. For \textit{DANNCE without BLUR}, we simply omit blurring from our DANNCE implementation. The results demonstrate two things: (1) blurring without image updates worsens performance against DANN and (2) DANNCE without blurring has no performance gain over DANN. This ablation tells us that while our performance gains are \textit{not} coming from blurring itself, blurring may be important for DANNCE on image datasets. This agrees with findings by Yosinski et al.~\cite{yosinskiunderstanding} which suggests blurring suppresses high frequency information produced by gradient updates so that image statistics do not differ too much from the real dataset.}

\centering 
\footnotesize
\setlength{\tabcolsep}{4pt}
\renewcommand{\arraystretch}{1.10}
\begin{tabular}{lcccc|cc}
\specialrule{.1em}{.1em}{0.1em} 
\textbf{PACS} & \textbf{art} & \textbf{cartoon} & \textbf{sketch} & \textbf{photo} & \textbf{avg} & \textbf{gain} \\ \hline
\multicolumn{7}{c}{\textbf{AlexNet}}                                                                                                   \\ \hline
DANN          & 71.2         & 72.1             & 66.3            & 88.3           & 74.5         & -           \\
DANN with BLUR          & 70.1         & 70.9             & 66.7            & 89.14           & 74.2         & -0.3           \\
DANNCE without BLUR          & 70.9         & 72.1             & 66.0            & 88.8           & 74.5         & 0.0           \\
DANNCE        & 70.9         & 72.0             & 67.9            & 89.6           & 75.1         & 0.6           \\
\specialrule{.1em}{.1em}{0.1em} 
\end{tabular}
\end{table}